\documentclass{article} 
\usepackage{iclr2027_conference,times}

\usepackage[utf8]{inputenc} 
\usepackage[T1]{fontenc}    
\usepackage{hyperref}       
\usepackage{url}            
\usepackage{booktabs}       
\usepackage{amsfonts}       
\usepackage{nicefrac}       
\usepackage{microtype}      
\usepackage[dvipsnames]{xcolor} 
\usepackage{amsmath,amsthm}
\usepackage{graphicx}
\usepackage{multirow}
\usepackage{algorithm}
\usepackage{algpseudocode}
\usepackage{soul}
\usepackage[table]{xcolor}
\usepackage{enumitem}
\usepackage{subcaption}
\usepackage{wrapfig}

\newcommand{\gc}[1]{\cellcolor{green!20} +#1}
\newcommand{\rc}[1]{\cellcolor{red!20} -#1}

\newcommand{\mymethod}{SIMBA}

\title{Rethinking Learning-Based Influence Maximization: Simple Neural Surrogates and Native Discrete Search}

\author{Yiqiao Liao \\
UC San Diego \\
\texttt{yil345@ucsd.edu} \\
\And
Parinaz Naghizadeh \\
UC San Diego\\
\texttt{parinaz@ucsd.edu}
}

\iclrfinalcopy 
\begin{document}

\maketitle

\begin{abstract}
Existing learning-based influence maximization frameworks rely heavily on complex neural architectures and continuous optimization over seed representations. We challenge this paradigm with \mymethod{}, a diffusion-model-agnostic framework pairing a lightweight neural surrogate with direct discrete search. \mymethod{} introduces three key components: 1) uniformly anchored node embeddings that eliminate initialization noise and encourage learning driven by graph topology and diffusion pattern, 2) a shallow two-layer graph neural network surrogate predicting final infection states, and 3) batched multi-swap simulated annealing that explores combinatorial seed space without gradients or continuous relaxation. By shifting compute from complex representation learning to effective discrete search, \mymethod{} drastically cuts time-to-solution while achieving superior influence spread and data efficiency. Our code is available at {\small\url{https://github.com/yl489/rethink-IM}}.\looseness=-1
\end{abstract}

\section{Introduction}

Influence Maximization (IM), the task of identifying a small subset of seed nodes that maximizes the expected spread of information over a network, is a fundamental problem in network science, viral marketing, and epidemiology~\citep{kempe2003maximizing, chen2010scalable, neophytou2024promoting}. Recent learning-based frameworks such as DeepIM~\citep{ling2023deepim} and DeepSN~\citep{hevapathige2025deepsn} demonstrate that influence spread can be predicted without explicit knowledge of the underlying diffusion model using only observed pairs of initial seed configurations and their corresponding diffusion outcomes. Once trained, these surrogates can replace expensive diffusion simulations, enabling diffusion-model-agnostic IM on large graphs with state-of-the-art performance. Despite this progress, existing neural surrogate frameworks rely on increasingly sophisticated architectures and optimization pipelines. They share a common assumption: effective diffusion-model-agnostic IM requires complex neural architectures together with continuous optimization over seed representations.

In this work, we challenge this assumption and propose \mymethod{}, which consists of three complementary components. First, we introduce \emph{uniformly anchored node embeddings}, where every learnable node embedding is initialized from a shared random anchor, and the input seed configuration is encoded by masking non-seed embeddings. This design removes initialization noise and allows message passing to construct node representations purely from graph topology and diffusion pattern. Second, we employ a \emph{lightweight neural surrogate} implemented using a two-layer graph neural network (GNN) that predicts the final infection state of every node given an initial seed configuration. Finally, we formulate IM as a discrete optimization problem and solve it using a \emph{batched multi-swap simulated annealing} algorithm that repeatedly queries the learned surrogate. Since candidate seed sets are evaluated in parallel using batched inference, the optimization process efficiently explores the discrete search space without requiring repeated diffusion simulations or gradient computations.\looseness=-1

An important consequence of this framework is a shift in computational effort from expensive representation learning to effective combinatorial search. Existing approaches invest substantial computation in learning expressive continuous representations to enable efficient gradient-based inference. In contrast, the lightweight surrogate employed by \mymethod{} converges rapidly, allowing significantly more computation to be devoted to high-quality discrete search. As a result, the overall time-to-solution is substantially reduced, while the output seed sets consistently yield significantly higher influence.\looseness=-1

Our experiments further reveal an unexpected degree of data efficiency. We find that a small fraction of seed and diffusion outcomes pairs used by existing neural surrogate frameworks is sufficient for \mymethod{} to learn an effective surrogate. The resulting reduction in training cost, together with more aggressive surrogate-guided search, allows the framework to scale efficiently to graphs containing hundreds of thousands of nodes while achieving state-of-the-art performance.

Finally, we investigate why such a lightweight surrogate generalizes effectively to previously unseen seed configurations. Through an analysis of the learned embedding space, we observe that uniformly anchored node embeddings often naturally evolve into a highly structured low-dimensional geometric manifold after learning. This emergent organization produces smooth surrogate predictions for novel combinations of seed nodes, providing a geometric explanation for the strong generalization performance observed during discrete optimization.

Our contributions are summarized as follows.
\begin{itemize}[nosep]
\item We propose \mymethod{}, a diffusion-model-agnostic IM framework that combines a lightweight GNN neural surrogate with batched multi-swap simulated annealing for direct optimization over the discrete seed space.
\item We introduce uniformly anchored node embeddings that enable seed configurations to be represented through embedding masking while encouraging representation learning driven by graph topology and diffusion pattern.
\item We demonstrate that combining a lightweight surrogate with direct discrete optimization outperforms existing continuous optimization frameworks, achieving higher influence spread with significantly lower time-to-solution.
\item We show that \mymethod{} remains highly data efficient, requiring only a small fraction of diffusion observations while maintaining strong performance on large-scale graphs.
\item We provide an empirical analysis of the learned embedding geometry, offering insight into why the proposed surrogate generalizes effectively to previously unseen seed configurations.
\end{itemize}

\section{Related Work}

Traditional IM methods typically offer principled optimization and provable approximation bounds but often require full access to the underlying diffusion model and are limited to specific diffusion models~\citep{kempe2003maximizing, leskovec2007cost, saito2012efficient, borgs2014maximizing, tang2014influence, tang2015imm, nguyen2016stop, li2019tiptop, hong2020efficient}. In empirical real-world networks, the underlying propagation rules are rarely known or easily parameterized, severely restricting the applicability of these approaches.

There is a growing literature proposing learning-based IM approaches. Most such approaches rely on reinforcement learning to train an agent that selects the next seed node~\citep{lin2015learning, li2019disco, tian2020deep, manchanda2020gcomb, li2022piano, chen2023touplegdd}. Since the reward is the marginal influence gain, these methods need to repeatedly observe the diffusion outcomes. There are also methods that directly train GNNs to predict the influence spread. \cite{panagopoulos2020IMINFECTOR} leverage fine-grained diffusion cascades as the training data; \cite{kumar2022influence} expect every node's influence as the training label. Addressing the performance limitations and scalability issues of all aforementioned methods, DeepIM~\citep{ling2023deepim} and DeepSN~\citep{hevapathige2025deepsn} utilize continuous optimization and sophisticated GNN architectures to train surrogate models that predict the influence spread of given seed sets and search for high-quality seed sets through gradient-based optimization. We provide a detailed review of DeepIM and DeepSN in Appendix~\ref{app:related}. Most recently, \cite{zhang2026imgnn} attempt to achieve cross-graph generalization; \cite{zhang2026dynaflux} incorporate more information into their model by observing the node states at each diffusion step. However, both these works require access to the underlying diffusion models to run simulations for model training.

\section{Problem formulation}

Let $G=(V, E)$ denote a directed graph with node set
$V=\{v_1,v_2,\dots,v_N\}$ and edge set $E$. The objective of IM is to identify a seed set $S\subset V$ with a fixed size $|S|=k$ that maximizes the expected information spread over the network. We denote the influence spread of a seed set by $\sigma(S)$, which represents the expected number of nodes activated under the underlying diffusion process. We consider a \textit{diffusion-model-agnostic} setting, where the objective is to learn the diffusion dynamics directly from observed diffusion outcomes without access to the underlying propagation mechanism.\looseness=-1

Specifically, given a graph $G$, we assume access to a training dataset $\mathcal{D} = \{(S_m,\mathbf y_m)\}_{m=1}^{M},$ where each seed set $S_m$ is represented by a binary indicator vector $\mathbf{s}_m\in\{0,1\}^{N}.$
The corresponding target $\mathbf y_m\in\{0,1\}^{N}$ denotes the observed final infection state after diffusion, where $y_{m,i}=1$ indicates that node $v_i$ is activated and $y_{m,i}=0$ otherwise. The goal of diffusion-model-agnostic IM is therefore to learn a surrogate mapping $f_\theta(G,\mathbf{s}) = \hat{\mathbf y},$ where $\hat{\mathbf y}\in\mathbb R^N$ approximates the final infection state produced by the unknown diffusion process. The predicted influence of a candidate seed set can then be estimated as $\hat{\sigma}(S) = \sum_{i=1}^{N}\hat y_i.$ Using this learned surrogate, the original IM problem becomes $S^* = \arg\max_{|S|=k} \hat{\sigma}(S),$ which can be solved without explicitly simulating the unknown diffusion process.\looseness=-1

\section{Method}\label{sec:method}

\subsection{Uniformly anchored node embeddings}
Given a graph $G=(V, E)$, we associate every node $v_i$ with a learnable embedding $\mathbf e_i\in\mathbb R^d$. This improves the expressiveness of the model but introduces the cold start problem on unseen nodes (i.e., $v_i\notin \cup_{m=1}^M S_m$). To resolve this, we introduce a simple uniform initialization anchor. Specifically, all node embeddings are initialized from the same random anchor: $\mathbf e_i=\mathbf c, \forall v_i\in V,$ where $\mathbf c\in\mathbb R^d$ is sampled once at initialization. By enforcing an identical initial coordinate across all node embeddings, we isolate the GNN from feature noise. Freed from initialization noise, the node embedding updates are driven by the graph topology and diffusion pattern. The anchor also acts as a special token that is easier for the model to interpret compared to unseen random embeddings. 

For a candidate seed set represented by a binary vector $\mathbf{s}\in\{0,1\}^{N},$ we construct the node feature matrix by masking the embeddings according to the seed configuration: $\mathbf{x}_i=s_i\mathbf e_i$ (i.e., $\mathbf{x}_i = \mathbf e_i$ if $v_i\in S$, and $\mathbf 0$ otherwise). This formulation allows the same learnable node representations to be reused across arbitrary seed configurations. The seed set is introduced only through masking, while the GNN learns how diffusion propagates from activated nodes through graph topology and diffusion outcomes.\looseness=-1

\subsection{Lightweight neural surrogate}
The goal of the neural surrogate is to approximate the unknown diffusion process by predicting the final infection state of every node from an initial seed configuration via a surrogate mapping $f_\theta(G,\mathbf{s})=\hat{\mathbf y}$. We instantiate $f_\theta$ using a lightweight two-layer GraphSAGE network~\citep{hamilton2017graphsage}. The input node features are the uniformly anchored node embeddings introduced in the previous section after masking according to the input seed configuration. Through neighborhood aggregation, the GNN propagates information from the activated seed nodes across the graph, allowing the network to approximate the underlying diffusion dynamics without requiring explicit knowledge of the diffusion model. We intentionally adopt a shallow and unconstrained architecture, as our objective is to learn a fast yet sufficiently accurate surrogate for efficient seed set search.

The surrogate is trained using the observation pairs in $\mathcal{D}$ by minimizing the mean squared error between the predicted and observed final infection states: $ \mathcal {L} (\theta) = \frac{1}{N} \sum_{i=1}^{N} (\hat y_i-y_i)^2$. After training, the surrogate parameters are frozen. For a candidate seed set $S$, the predicted influence is computed by aggregating the predicted node states, $\hat{\sigma}(S) = \sum_{i=1}^{N}\hat y_i$, which serves as the objective function for the downstream optimization algorithm. Consequently, seed set optimization requires only repeated forward passes through the lightweight surrogate, eliminating the need for gradient-based optimization, latent seed representations, or repeated diffusion simulations during inference.

\begin{algorithm}[t]
\caption{Batched Multi-Swap Simulated Annealing Guided by Neural Surrogate}
\label{alg:batch_sa}
\begin{algorithmic}[1]
\Require Trained surrogate $f_\theta$, graph $G$, seed budget $k$, number of steps $I$, batch size $B$, number of swaps $r$, initial temperature $T_0$, cooling rate $\alpha$

\State Initialize $B$ random seed sets $\{S_1,\dots,S_B\}$ with $|S_b|=k$
\State Evaluate $\hat{\sigma}(S_b)$ for all $b$ using one batched surrogate inference
\State Initialize $S^*$ as the highest-scoring candidate
\State $T\leftarrow\alpha T_0$

\For{$i=1,\dots,I$}
    \For{each candidate $S_b$}
        \State Sample $r$ nodes $R_b\subset S_b$ to remove
        \State Sample $r$ nodes $A_b\subset V\setminus S_b$ to add
        \State Construct neighbor:
        $
        S'_b=(S_b\setminus R_b)\cup A_b
        $
    \EndFor

    \State Evaluate all neighbors using $f_\theta$ with batched surrogate inference
    \For{each candidate $S_b$}
        \State Compute
        $
        \Delta_b=\hat{\sigma}(S'_b)-\hat{\sigma}(S_b)
        $
        \If{$\Delta_b>0$ or $\mathrm{rand}()<\exp(\Delta_b/T)$}
            \State Accept $S'_b$ as the new candidate
        \EndIf
    \EndFor

    \State Update $S^*$ with the best candidate found so far
    \State $T\leftarrow\alpha T$
\EndFor

\State \Return $S^*$
\end{algorithmic}
\end{algorithm}

\subsection{Batched multi-swap simulated annealing}
After training the surrogate, IM is formulated as the discrete optimization problem $S^* = \arg\max_{|S|=k}\hat{\sigma}(S).$ To solve this problem efficiently, we introduce a batched multi-swap simulated annealing procedure. Instead of optimizing a single seed set trajectory, we simultaneously maintain a batch of $B$ candidate solutions: $\mathcal{S} = \{S_1, S_2, \dots, S_B\}$. Each candidate satisfies the cardinality constraint $|S_b|=k.$ At each iteration, a neighboring solution is generated for every candidate by replacing $r$ currently selected nodes with $r$ unselected nodes. Specifically, for candidate $S_b$, we sample $R_b\subset S_b, A_b\subset V\setminus S_b,$ with $|R_b|=|A_b|=r,$ and construct $S'_b=(S_b\setminus R_b)\cup A_b.$ Because the numbers of removals and additions are identical, every candidate remains a valid size-$k$ seed set. All neighboring solutions are evaluated simultaneously through a single batched forward pass: $\hat{\sigma}(S'_1),\dots,\hat{\sigma}(S'_B) = f_\theta(\mathcal{S}').$ For each trajectory, we compute $\Delta_b = \hat{\sigma}(S'_b) - \hat{\sigma}(S_b).$ The proposed solution is accepted according to the simulated annealing rule: 
$p_{\mathrm{acc}} = 1$ if $\Delta_b > 0$, and $\exp(\Delta_b/T)$ otherwise, where $T$ is the current temperature. The temperature is updated after each iteration using geometric cooling: $T_{i+1}=\alpha T_i,$ where $\alpha$ is the cooling rate. Throughout optimization, we maintain the best solution observed across all parallel search trajectories. Algorithm~\ref{alg:batch_sa} summarizes the proposed optimization procedure.\looseness=-1

\subsection{Computational efficiency}
The computational philosophy of \mymethod{} differs fundamentally from existing learning-based IM frameworks. Previous approaches invest the majority of their computational budget in offline surrogate learning, employing complex neural architectures and continuous latent optimization to minimize the cost of seed set optimization during inference. In contrast, \mymethod{} intentionally redistributes computation by learning a lightweight neural surrogate and allocating the majority of the computational budget to surrogate-guided discrete search during inference. This design substantially reduces training overhead while enabling more extensive exploration of the combinatorial seed space.

From a computational perspective, both the surrogate learning stage and the discrete optimization stage scale linearly with the size of the graph. To further improve inference efficiency, \mymethod{} employs two complementary strategies. First, candidate seed sets are evaluated in parallel using batched inference, allowing multiple search trajectories to share a single forward pass through the surrogate. Second, the proposed multi-swap neighborhood operator explores larger regions of the discrete search space at each iteration, reducing the number of optimization steps required to reach high-quality solutions compared with conventional single-swap local search.

Additionally, we uncover and exploit an unexpected data efficiency property of diffusion-model-agnostic IM. As validated in Section~\ref{sec:data_efficiency}, the surrogate can be trained using only a small fraction of the available diffusion observations while maintaining comparable IM performance. This substantially reduces surrogate training time and enables more computational resources to be devoted to discrete optimization. On very large graphs, where both training and optimization become increasingly expensive, this data efficiency and more exploratory search lead to significant reductions in time-to-solution while preserving the quality of the discovered seed sets.

\section{Experiments}\label{sec:exp}

\textbf{Datasets.}
We evaluate \mymethod{} on five real-world benchmark graphs, Jazz, Network Science (NS), Power Grid (PG)~\citep{rossi2015network}, Cora-ML~\citep{mccallum2000automating}, and  Digg~\citep{lerman2012social}, and one synthetic Erdős–Rényi graph~\citep{ling2023deepim} spanning a wide range of network sizes and structural characteristics. We consider the linear threshold (LT)~\citep{granovetter1978threshold}, independent cascade (IC)~\citep{kempe2003maximizing}, and susceptible-infected-susceptible~\citep{kermack1927epidemics} (SIS) diffusion models. Dataset details are provided in Appendix~\ref{app:datasets}.

\textbf{Baselines.}
We compare against \textit{traditional IM} methods, including IMM~\citep{tang2015imm}, OPIM~\citep{tang2018opim}, and SubSIM~\citep{guo2020subsim}, an \textit{online IM} method, OIM~\citep{lei2015oim}, and \textit{learning-based IM} methods, including IMINFECTOR~\citep{panagopoulos2020IMINFECTOR}, PIANO~\citep{li2022piano}, ToupleGDD~\citep{chen2023touplegdd}, DeepIM~\citep{ling2023deepim} and DeepSN~\citep{hevapathige2025deepsn}. Additionally, we include DynaFLUX~\citep{zhang2026dynaflux} in Table~\ref{tab:dynaflux} in Appendix~\ref{app:dynaflux}.

\textbf{Evaluation metrics.}
Following prior works, the primary evaluation metric is the achieved influence spread under the underlying diffusion process with 1\%, 5\%, 10\%, and 20\% of the nodes as seed nodes. We provide a note on how the reported spread is computed in Appendix~\ref{app:eval}. We additionally report the time-to-solution (TTS), defined as the combined surrogate training time and surrogate-guided search time required to produce the final seed set.

\textbf{Implementation details.} Given the number of sources of randomness and the probabilistic nature of the components in the pipeline (data split, model and search initializations, and diffusion simulation), we set the initial temperature $T_0 = 1e^{-6}$ and the cooling rate $\alpha = 1$ for all experiments included in the main paper to avoid further conflating with the effect of random candidate solution acceptance. We report the results with a different set of temperature and cooling parameters in Appendix~\ref{app:temperature}. Additional implementation details are provided in Appendix~\ref{app:impl_details}.

\begin{table}[t]
    \centering
    \caption{Influence spread (\%) under the IC, LT, and SIS models. The best (resp. second best) results are in bold (resp. underlined). $\Delta$ is the difference between \mymethod{} and the best baseline.}
    \vspace{-3.5mm}
    \label{tab:im_results}
    \resizebox{\textwidth}{!}{
    \begingroup \setlength{\tabcolsep}{3pt}
    \begin{tabular}{@{}lcccccccccccccccccccccccc}
    \toprule
        & \multicolumn{4}{c}{Cora-ML (IC)} & \multicolumn{4}{c}{Network Science (IC)} & \multicolumn{4}{c}{Power Grid (IC)} & \multicolumn{4}{c}{Jazz (IC)} & \multicolumn{4}{c}{Synthetic (IC)} & \multicolumn{4}{c}{Digg (IC)} \\
        \cmidrule(lr){2-5} \cmidrule(lr){6-9} \cmidrule(lr){10-13} \cmidrule(lr){14-17} \cmidrule(lr){18-21} \cmidrule(lr){22-25}
        Method & 1\% & 5\% & 10\% & 20\% & 1\% & 5\% & 10\% & 20\% & 1\% & 5\% & 10\% & 20\% & 1\% & 5\% & 10\% & 20\% & 1\% & 5\% & 10\% & 20\% & 1\% & 5\% & 10\% & 20\% \\
        \midrule
        IMM        &  8.1 & 26.2 & 37.3 & 50.2 &  5.2 & 16.8 & 27.0 & 45.7 &  4.3 & 17.4 & 31.5 & 51.1 &  2.6 & 20.1 & 31.4 & 42.8 &  9.2 & 26.2 & 36.3 & 51.6 &  7.4 & 18.4 & 32.8 & 49.6 \\
        OPIM       & 13.4 & 26.9 & 37.4 & 50.9 &  6.6 & 19.4 & 28.9 & 48.6 &  5.7 & 17.7 & 29.7 & 50.1 &  2.4 & 20.1 & 34.4 & 46.8 &  9.6 & 25.3 & 36.6 & 51.7 &  7.6 & 18.5 & 32.9 & 48.9 \\
        SubSIM     & 10.1 & 25.7 & 36.8 & 51.1 &  4.8 & 15.4 & 27.9 & 44.8 &  4.6 & 19.2 & 31.7 & 50.2 &  3.6 & 18.8 & 37.6 & 44.7 &  9.5 & 26.7 & 36.5 & 51.5 &  7.5 & 18.9 & 33.3 & 49.4 \\
        \midrule
        OIM        & 8.9 & 27.6 & 38.0 & 51.3 & 4.2 & 16.7 & 26.5 & 48.2 & 5.7 & 17.5 & 31.9 & 50.8 & 2.0 & 18.5 & 36.3 & 42.2 & 9.6 & 26.2 & 36.7 & 51.3 & 7.8 & 18.2 & 33.1 & 49.6 \\
        \midrule
        IMINFECTOR &  9.6 & 26.8 & 37.7 & 50.6 &  5.4 & 17.9 & 27.8 & 47.6 &  5.4 & 18.2 & 31.6 & 50.9 &  3.6 & 19.7 & 37.5 & 45.9 &  9.1 & 26.2 & 36.1 & 51.5 &  7.9 & 18.6 & 33.5 & 49.8 \\
        PIANO      &  9.8 & 25.2 & 37.4 & 51.1 &  4.7 & 16.3 & 27.1 & 47.2 &  5.3 & 18.1 & 31.7 & 50.2 &  2.2 & 19.2 & 36.6 & 43.2 &  9.1 & 26.4 & 36.2 & 51.6 &  --  &  --  &  --  &  --  \\
        ToupleGDD  & 10.6 & 27.5 & 38.5 & 51.5 &  6.3 & 17.8 & 28.3 & 50.5 &  5.4 & 19.3 & 31.6 & 51.3 &  3.3 & 20.4 & 37.2 & 45.7 &  9.5 & 26.8 & 37.1 & 51.4 &  --  &  --  &  --  &  --  \\
        \midrule
        DeepIM     & 14.1 & 28.1 & 39.6 & 52.4 &  7.8 & 20.9 & 31.5 & 51.2 &  6.3 & 21.0 & 32.5 & 52.4 &  4.9 & 23.3 & \underline{41.5} & 49.9 & 11.6 & 27.4 & 38.7 & 52.1 &  8.4 & 19.3 & 34.2 & 51.3 \\
        DeepSN     & 11.5 & 25.6 & 40.9 & 52.8 &  6.2 & 22.0 & 32.0 & 52.4 &  6.4 & 24.0 & 36.9 & \textbf{61.0} &  8.5 & 26.9 & \textbf{41.6} & \textbf{53.8} & 10.6 & 27.8 & \underline{38.8} & \underline{52.8} &  8.9 & 19.5 & 35.2 & 52.8 \\
        \midrule
        \mymethod{} (C) & 14.2 & 33.2 & 39.8 & 54.7 & 9.5 & \underline{24.0} & \underline{36.6} & 50.3 & \textbf{8.5} & \underline{26.7} & \underline{40.5} & 57.6 & 10.4 & \underline{27.2} & 36.7 & 50.0 & \underline{11.9} & \underline{28.3} & 38.5 & 52.2 & \underline{60.3} & 66.3 & \underline{72.3} & \underline{80.6} \\
        \mymethod{} (U) & \underline{16.9} & \underline{33.3} & \underline{43.1} & \underline{57.3} & \textbf{9.7} & \underline{24.0} & \underline{36.6} & \underline{57.8} & \underline{8.2} & 26.2 & 40.4 & 59.5 & \textbf{12.3} & \textbf{27.6} & 37.5 & \underline{51.1} & 11.6 & \textbf{28.4} & \textbf{39.4} & \textbf{53.8} & \textbf{60.4} & \textbf{66.5} & \underline{72.3} & \underline{80.6} \\
        \mymethod{} (U*) & \textbf{17.2} & \textbf{33.6} & \textbf{43.3} & \textbf{57.4} & \underline{9.6} & \textbf{24.4} & \textbf{37.1} & \textbf{58.1} & \textbf{8.5} & \textbf{26.9} & \textbf{41.3} & \underline{60.1} & \underline{11.7} & 26.7 & 37.5 & 50.8 & \textbf{12.0} & \underline{28.3} & \textbf{39.4} & \textbf{53.8} & \underline{60.3} & \underline{66.4} & \textbf{72.5} & \textbf{81.2} \\
        \midrule
        $\Delta$ & \gc{3.1} & \gc{5.5} & \gc{2.4} & \gc{4.6} & \gc{1.9} & \gc{2.4} & \gc{5.1} & \gc{5.7} & \gc{2.1} & \gc{2.9} & \gc{4.4} & \rc{0.9} & \gc{3.8} & \gc{0.7} & \rc{4.1} & \rc{2.7} & \gc{0.4} & \gc{0.6} & \gc{0.6} & \gc{1.0} & \gc{51.5} & \gc{47.0} & \gc{37.3} & \gc{28.4} \\
    \bottomrule
    \end{tabular}\endgroup}
    
    \resizebox{\textwidth}{!}{
    \begingroup \setlength{\tabcolsep}{2.5pt}
    \begin{tabular}{@{}lcccccccccccccccccccccccc}
        \toprule
        & \multicolumn{4}{c}{Cora-ML (LT)} & \multicolumn{4}{c}{Network Science (LT)} & \multicolumn{4}{c}{Power Grid (LT)} & \multicolumn{4}{c}{Jazz (LT)} & \multicolumn{4}{c}{Synthetic (LT)} & \multicolumn{4}{c}{Digg (LT)} \\
        \cmidrule(lr){2-5} \cmidrule(lr){6-9} \cmidrule(lr){10-13} \cmidrule(lr){14-17} \cmidrule(lr){18-21} \cmidrule(lr){22-25}
        Method & 1\% & 5\% & 10\% & 20\% & 1\% & 5\% & 10\% & 20\% & 1\% & 5\% & 10\% & 20\% & 1\% & 5\% & 10\% & 20\% & 1\% & 5\% & 10\% & 20\% & 1\% & 5\% & 10\% & 20\% \\
        \midrule
        IMM        & 1.7 & 34.8 & 52.2 & 66.4 & 2.5 & 11.9 & 18.1 & 33.6 & 4.6 & 19.9 & 31.7 & 56.9 & 1.4 & 5.7 & 13.4 & 24.5 & 1.1 & 5.2 & 13.1 & 66.9 & 2.4 & 10.8 & 37.4 & 55.6 \\
        OPIM       & 2.3 & 36.9 & 51.2 & 71.5 & 1.6 & 12.0 & 18.8 & 34.1 & 4.4 & 21.6 & 29.4 & 55.5 & 1.4 & 6.9 & 12.6 & 20.9 & 1.3 & 5.2 & 12.6 & 62.1 & 2.1 & 11.3 & 38.2 & 57.1 \\
        SubSIM     & 1.7 & 33.6 & 54.7 & 70.1 & 1.8 & 10.4 & 19.2 & 34.1 & 4.5 & 21.1 & 31.2 & 57.4 & 1.4 & 5.9 & 11.4 & 21.2 & 1.4 & 5.5 & 13.1 & 69.6 & 2.4 & 11.3 & 37.9 & 56.9 \\
        \midrule
        IMINFECTOR & 2.1 & 33.9 & 51.3 & 70.6 & 2.1 & 11.8 & 18.7 & 34.5 & 4.2 & 21.3 & 31.6 & 56.2 & 1.4 & 6.2 & 13.5 & 22.8 & 1.3 & 5.2 & 12.9 & 67.4 & 2.2 & 11.1 & 38.9 & 58.7 \\
        PIANO      & 2.1 & 33.5 & 53.3 & 69.8 & 2.1 & 11.3 & 19.1 & 33.9 & 4.3 & 21.3 & 31.4 & 57.1 & 1.1 & 6.2 & 12.1 & 22.4 & 1.2 & 5.2 & 12.9 & 67.4 & -- & -- & -- & -- \\
        ToupleGDD  & 2.3 & 36.2 & 54.5 & 70.9 & 2.8 & 12.4 & 19.8 & 34.6 & 4.8 & 21.9 & 32.6 & 58.1 & 1.4 & 6.5 & 12.9 & 23.6 & 1.3 & 5.5 & 13.4 & 70.2 & -- & -- & -- & -- \\
        \midrule
        DeepIM     & 13.4 & 69.2 & 83.5 & 94.1 & 4.1 & 16.6 & 26.7 & 41.5 & 6.3 & 24.4 & 46.8 & 71.7 & \underline{1.9} & 6.5 & 16.4 & \underline{99.1} & \underline{1.5} & 6.5 & 15.5 & \underline{99.9} & 3.5 & 15.9 & 41.3 & 76.2 \\
        DeepSN     & 7.4 & 40.7 & 68.2 & 95.3 & 2.9 & 14.4 & 25.3 & 52.0 & 6.3 & 24.6 & 47.2 & 73.2 & \textbf{2.0} & 6.7 & 14.9 & 96.9 & \textbf{1.6} & 5.8 & 13.5 & \underline{99.9} & 3.2 & 16.1 & 41.7 & 72.1 \\
        \midrule
        \mymethod{} (C) & \textbf{20.7} & 76.7 & 84.8 & 89.4 & 4.9 & 22.1 & 28.6 & 41.9 & \textbf{7.8} & \textbf{32.5} & 54.2 & 77.4 & 1.3 & 7.4 & \textbf{71.2} & 98.6 & 1.1 & 6.5 & \underline{15.9} & \textbf{100.0} & 12.3 & 46.4 & 85.0 & 90.6 \\
        \mymethod{} (U) & \underline{18.9} & \underline{80.0} & \textbf{93.3} & \underline{98.8} & \underline{6.8} & \textbf{26.5} & \textbf{41.7} & \underline{66.4} & \underline{7.6} & 31.0 & \underline{58.4} & \underline{88.1} & \textbf{2.0} & \textbf{30.4} & 48.0 & \textbf{100.0} & 1.4 & \underline{7.0} & \underline{15.9} & \textbf{100.0} & \underline{15.4} & \underline{81.9} & \underline{85.2} & \underline{90.9} \\
        \mymethod{} (U*) & 18.0 & \textbf{80.3} & \underline{93.0} & \textbf{98.9} & \textbf{7.0} & \underline{26.3} & \underline{41.6} & \textbf{66.6} & \textbf{7.8} & \underline{32.4} & \textbf{59.5} & \textbf{88.7} & \textbf{2.0} & \underline{22.5} & \underline{58.5} & \textbf{100.0} & 1.4 & \textbf{7.1} & \textbf{16.4} & \textbf{100.0} & \textbf{18.6} & \textbf{82.4} & \textbf{86.3} & \textbf{91.8} \\
        \midrule
        $\Delta$ & \gc{7.3} & \gc{11.1} & \gc{9.8} & \gc{3.6} & \gc{2.9} & \gc{9.9} & \gc{15.0} & \gc{14.6} & \gc{1.5} & \gc{7.9} & \gc{12.3} & \gc{15.5} & 0.0 & \gc{23.5} & \gc{54.8} & \gc{0.9} & \rc{0.2} & \gc{0.6} & \gc{0.9} & \gc{0.1} & \gc{15.1} & \gc{66.3} & \gc{44.6} & \gc{15.6} \\
        \bottomrule
    \end{tabular}\endgroup}

    \resizebox{\textwidth}{!}{
    \begingroup \setlength{\tabcolsep}{4pt}
    \begin{tabular}{@{}lcccccccccccccccccccccccc}
        \toprule
        & \multicolumn{4}{c}{Cora-ML (SIS)} & \multicolumn{4}{c}{Network Science (SIS)} & \multicolumn{4}{c}{Power Grid (SIS)} & \multicolumn{4}{c}{Jazz (SIS)} & \multicolumn{4}{c}{Synthetic (SIS)} & \multicolumn{4}{c}{Digg (SIS)} \\
        \cmidrule(lr){2-5} \cmidrule(lr){6-9} \cmidrule(lr){10-13} \cmidrule(lr){14-17} \cmidrule(lr){18-21} \cmidrule(lr){22-25}
        Method & 1\% & 5\% & 10\% & 20\% & 1\% & 5\% & 10\% & 20\% & 1\% & 5\% & 10\% & 20\% & 1\% & 5\% & 10\% & 20\% & 1\% & 5\% & 10\% & 20\% & 1\% & 5\% & 10\% & 20\% \\
        \midrule
        IMM        & 2.0 & 9.5 & 15.4 & 27.6 & 1.3 & 5.6 & 12.2 & 22.1 & 1.1 & 5.6 & 11.0 & 22.9 & 7.6 & 37.8 & 55.6 & 67.1  & 2.7 & 12.6 & 20.9 & 37.3 & 2.5 & 9.4 & 16.3 & 32.6 \\
        OPIM       & 2.3 & 9.3 & 16.2 & 27.2 & 1.4 & 5.9 & 13.0 & 22.1 & 1.2 & 5.9 & 11.2 & 22.4 & 5.7 & 44.7 & 58.6 & 68.3 & 2.8 & 12.5 & 20.2 & 36.1 & 2.3 & 9.3 & 16.5 & 32.1 \\
        SubSIM     & 2.3 & 9.2 & 16.9 & 28.8 & 1.5 & 5.6 & 12.2 & 23.3 & 1.2 & 5.6 & 11.4 & 21.9 & 2.9 & 30.1 & 53.8 & 67.0 & 2.5 & 12.6 & 20.2 & 36.5 & 2.5 & 9.5 & 16.1 & 32.3 \\
        IMINFECTOR & 2.1 & 9.4 & 16.1 & 27.9 & 1.7 & 5.8 & 12.4 & 22.3 & \underline{1.3} & 5.5 & 12.4 & 23.1 & 8.8 & 35.4 & 54.8 & 66.2 & 2.5 & 12.4 & 20.5 & 36.2 & 2.3 & 9.1 & 16.4 & 32.4 \\
        \midrule
        DeepIM     & \underline{7.1} & \textbf{16.1} & \underline{21.9} & 30.8 & \textbf{2.7} & 8.7 & \underline{15.1} & \underline{25.1} & \textbf{1.9} & \underline{7.6} & 13.3 & 23.8 & 27.1 & \textbf{57.1} & \textbf{68.1} & \textbf{74.1} & 3.2 & \underline{14.4} & 24.5 & 39.1 & \underline{5.6} & 11.4 & \underline{18.8} & \textbf{36.3} \\
        \midrule
        \mymethod{} (C) & \textbf{7.2} & 15.7 & \underline{21.9} & 31.1 & \underline{2.5} & \underline{8.8} & 14.6 & 23.8 & \textbf{1.9} & \textbf{7.8} & \textbf{14.1} & 24.7 & \textbf{33.6} & 55.4 & 63.3 & 70.3 & \textbf{3.5} & \textbf{14.6} & \underline{24.8} & 39.7 & \textbf{16.5} & \underline{20.3} & \textbf{24.9} & 33.5 \\
        \mymethod{} (U) & \textbf{7.2} & 15.7 & \textbf{22.5} & \textbf{32.5} & 2.4 & \underline{8.8} & \textbf{15.3} & \textbf{26.5} & \textbf{1.9} & \textbf{7.8} & \underline{14.0} & \underline{25.1} & 30.8 & \underline{56.7} & \underline{64.7} & \underline{71.6} & \underline{3.4} & 14.3 & 24.7 & \underline{40.2} & \textbf{16.5} & \underline{20.3} & \textbf{24.9} & 33.5 \\
        \mymethod{} (U*) & \textbf{7.2} & \underline{15.8} & \textbf{22.5} & \underline{32.4} & 2.4 & \textbf{8.9} & \textbf{15.3} & \textbf{26.5} & \textbf{1.9} & \textbf{7.8} & \textbf{14.1} & \textbf{25.2} & \underline{32.1} & 56.6 & 64.3 & 71.2 & \textbf{3.5} & \textbf{14.6} & \textbf{24.9} & \textbf{40.4} & \textbf{16.5} & \textbf{20.4} & \textbf{24.9} & \underline{33.6} \\
        \midrule
        $\Delta$ & \gc{0.1} & \rc{0.3} & \gc{0.6} & \gc{1.7} & \rc{0.2} & \gc{0.2} & \gc{0.2} & \gc{1.4} & 0.0 & \gc{0.2} & \gc{0.8} & \gc{1.4} & \gc{6.5} & \rc{0.4} & \rc{3.4} & \rc{2.5} & \gc{0.3} & \gc{0.2} & \gc{0.4} & \gc{1.3} & \gc{10.9} & \gc{9.0} & \gc{6.1} & \rc{2.7} \\
        \bottomrule
    \end{tabular}\endgroup}
\end{table}

\subsection{Influence maxmization performance}

We first evaluate whether the proposed framework is capable of identifying high-quality seed sets despite employing a lightweight neural surrogate and direct discrete optimization.
We consider three \mymethod{} variants: 1) the constrained setting (C) where the search is only done over nodes seen during surrogate training, 2) the unconstrained setting (U) where the search space equals the entire node set, and 3) the two-stage constrained-to-unconstrained setting (U*) where the search is first done over the seen nodes, and the resulting seed sets are used as the initializations for the following unconstrained search.
Table~\ref{tab:im_results} compares the influence spread obtained by \mymethod{} across all benchmark datasets. As noted by \cite{zhang2026dynaflux}, the evaluation protocol of DeepSN\footnote{https://github.com/Aselahp/DeepSN.} for the SIS experiments is different from that of other baselines: it runs 200 steps of simulation as opposed to 100. Thus, we also exclude DeepSN from comparison for the SIS experiments.

Overall, \mymethod{} consistently achieves the strongest performance, obtaining the highest influence spread in 62 of the 72 (86.1\%) settings. The improvements are frequently substantial rather than marginal, indicating that a lightweight neural surrogate coupled with direct discrete search can outperform considerably more sophisticated learning-based frameworks. These results suggest that the primary limitation of existing learning-based approaches is not the expressive power of the surrogate model, but rather the optimization strategy used to identify seed sets, and that neither highly expressive GNN architectures nor continuous optimization are prerequisites for effective IM. Although \mymethod{} does not achieve the best result in every setting, the remaining cases exhibit relatively small performance differences, indicating that the proposed framework remains highly competitive across diverse graph structures and diffusion scenarios.

Moreover, expanding the search space consistently improves IM performance. The unconstrained strategy (U) is no worse than the constrained setting (C) in 58 of the 72 (80.6\%) settings, meaning that the learned surrogate generalizes effectively to previously unseen nodes. If the surrogate failed to produce reliable predictions outside the seen nodes, enlarging the search space would introduce noisy objective evaluations and degrade optimization performance. Instead, the consistent improvements indicate that the surrogate successfully extrapolates beyond the node configurations encountered during training. The two-stage strategy (U*) further strengthens this claim. By first identifying a high-quality solution within the constrained search space and subsequently refining it over the full node set, U* is no worse than C in 66 of the 72 (91.7\%) settings. Since every refinement step is guided solely by the surrogate's predictions, the consistent improvement establishes that the surrogate provides sufficiently reliable evaluations of previously unseen seed configurations to guide effective optimization. These results provide strong empirical evidence that \mymethod{} achieves robust generalization, allowing the surrogate to meaningfully evaluate seed sets containing nodes that were never explored during training.\looseness=-1

\subsection{Time-to-solution}

\begin{wrapfigure}{r}{0.39\linewidth}
    \vspace{-4mm}
    \centering
    \includegraphics[width=\linewidth]{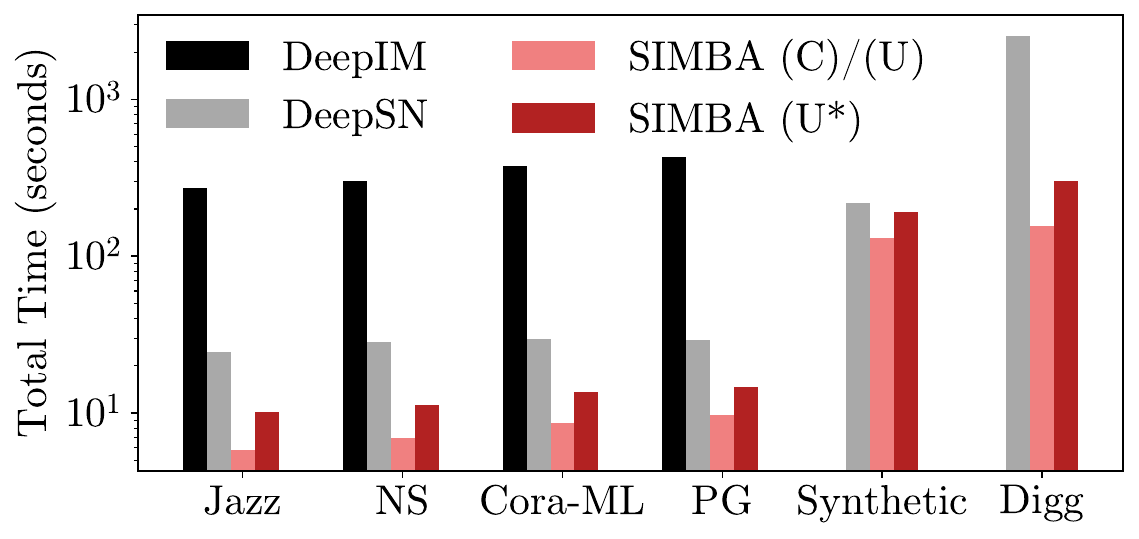}
    \vspace{-7mm}
    \caption{TTS across datasets.}
    \vspace{-7mm}
    \label{fig:runtime}
\end{wrapfigure}

High influence quality alone is insufficient for practical deployment. Existing learning-based IM frameworks often achieve fast inference only after substantial offline computation devoted to training. Consequently, reporting inference time alone provides an incomplete picture of the overall computational cost. As DeepIM runs much faster than earlier learning-based frameworks~\citep{ling2023deepim}, our runtime analysis focuses on DeepIM\footnote{https://github.com/triplej0079/DeepIM.}, DeepSN, and \mymethod{}.\looseness=-1

Table~\ref{tab:runtime} reports the TTS breakdown, and Figure~\ref{fig:runtime} compares the TTS of all approaches. Despite allocating more computation to inference for discrete optimization, \mymethod{} achieves significantly lower TTS because its lightweight surrogate trains rapidly and each optimization iteration requires only a batched forward pass through the fixed network. 
However, as shown in Table~\ref{tab:runtime}, as the graph size increases (from Jazz to Synthetic), the training time of \mymethod{} grows much faster than the inference time and gradually takes over. Additionally, because the search space grows exponentially with respect to the graph size, a large number of search steps is necessary to find high-quality seed sets on large graphs. Due to these two reasons, the TTS advantage of \mymethod{} over DeepSN is diminishing.\looseness=-1

\begin{table}[t]
    \centering
    \caption{TTS breakdown (trainining/inference).}
    \vspace{-3.5mm}
    \label{tab:runtime}
    \resizebox{0.9\textwidth}{!}{
    \begin{tabular}{lllllll}
        \toprule
        & Jazz & Network Science & Cora-ML & Power Grid & Synthetic & Digg \\
        \midrule
        \multirow{2}{*}{DeepIM}  
        & 274.33s & 300.58s & 376.69s & 426.59s & \multicolumn{1}{c}{\multirow{2}{*}{--}} & \multicolumn{1}{c}{\multirow{2}{*}{--}} \\
        & (272.64s/1.69s) & (298.80s/1.78s) & (374.80s/1.89s) & (424.70s/1.89s) & & \\
        \midrule
        \multirow{2}{*}{DeepSN}  
        & 24.62s & 28.44s & 29.50s & 29.01s & 219.19s & 2209.49s \\
        & (23.54s/1.09s) & (24.49s/3.95s) & (28.02s/1.49s) & (27.28s/1.73s) & (211.40s/7.79s) & (2243.58s/308.56s) \\
        \midrule
        \multirow{2}{*}{\mymethod{} (C)/(U)}
        & 5.81s & 6.89s & 8.62s & 9.78s & 130.80s & \textit{154.86s} \\
        & (0.72s/5.09s) & (1.67s/5.22s) & (2.91s/5.71s) & (4.06s/5.72s) & (69.77s/61.03s) & (\textit{8.91s}/\textit{145.95s}) \\
        \midrule
        \multirow{2}{*}{\mymethod{} (U*)}
        & 10.09s & 11.32s & 13.52s & 14.71s & 190.96s  & \textit{299.85s} \\
        & (0.72s/9.37s) & (1.67s/9.65s) & (2.91s/10.62s) & (4.06s/10.65s) & (69.77s/121.19s) & (\textit{8.91s}/\textit{290.94s}) \\
        \bottomrule
    \end{tabular}}
\end{table}

\begin{wrapfigure}{r}{0.39\linewidth}
    \vspace{-4mm}
    \centering
    \includegraphics[width=\linewidth]{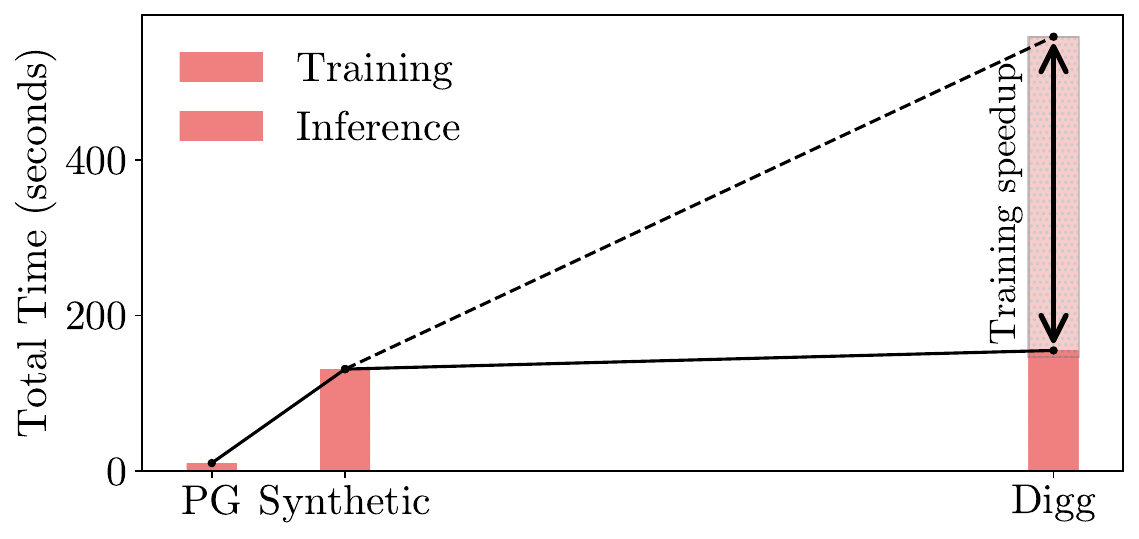}
    \vspace{-7mm}
    \caption{TTS scalability of \mymethod{}.}
    \vspace{-3mm}
    \label{fig:runtime_scalability}
\end{wrapfigure}

To improve the TTS scalability of \mymethod{}, we implement two strategies for experiments on Digg that target the training and inference stages, respectively. First, we reduce the number of training samples from 80 to 2. Second, we reduce the number of search steps from 50,000 to 20,000 and increase the number of swaps from 1 to 20 to enable more exploratory search. Table~\ref{tab:digg} in Appendix~\ref{app:digg} compares the influence spread achieved by the two search parameter settings. The resulting TTS of \mymethod{} on Digg is up to an order of magnitude faster than DeepSN and can remain comparable to that on Synthetic, as shown in Table~\ref{tab:runtime}, highlighting the exceptional TTS scalability of \mymethod{}, which is further illustrated in Figure~\ref{fig:runtime_scalability}.\looseness=-1

\begin{table}[t]
    \centering
    \caption{Influence spread (\%) achieved by \mymethod{} with varying fractions of training data.}
    \vspace{-3.5mm}
    \label{tab:data_efficiency}
    \resizebox{\textwidth}{!}{
    \begin{tabular}{c|lcccccccccccccccc@{}}
    \toprule
        \multicolumn{1}{l}{} & & \multicolumn{4}{c}{Cora-ML (IC)} & \multicolumn{4}{c}{Network Science (IC)} & \multicolumn{4}{c}{Power Grid (IC)} & \multicolumn{4}{c}{Jazz (IC)} \\
        \cmidrule(lr){3-6} \cmidrule(lr){7-10} \cmidrule(lr){11-14} \cmidrule(lr){15-18}
        \multicolumn{1}{c}{Training data} & Method & 1\% & 5\% & 10\% & 20\% & 1\% & 5\% & 10\% & 20\% & 1\% & 5\% & 10\% & 20\% & 1\% & 5\% & 10\% & 20\% \\
        \midrule
        1\% & \mymethod{} (U) & 16.8 & 32.2 & 42.8 & 57.2 & 9.7 & 23.8 & 37.0 & 58.0 & 7.9 & 25.8 & 40.2 & 59.3 & 12.1 & 27.5 & 37.5 & 49.3 \\
        \midrule
        5\% & \mymethod{} (U) &  16.9 & 32.5 & 43.2 & 56.8 & 9.7 & 24.0 & 36.8 & 58.0 & 8.2 & 25.8 & 40.4 & 59.5 & 12.3 & 27.5 & 37.1 & 50.2 \\
        \midrule
        20\% & \mymethod{} (U) & 16.9 & 32.4 & 43.1 & 57.0 & 9.7 & 23.9 & 37.1 & 58.0 & 7.9 & 25.9 & 40.6 & 59.4 & 11.7 & 26.6 & 36.3 & 50.3 \\
        \midrule
        100\% & \mymethod{} (U) & 16.9 & 33.3 & 43.1 & 57.3 & 9.7 & 24.0 & 36.6 & 57.8 & 8.2 & 26.2 & 40.4 & 59.5 & 12.3 & 27.6 & 37.5 & 51.1 \\
    \bottomrule
    \end{tabular}}
    
    \resizebox{\textwidth}{!}{
    \begin{tabular}{c|lcccccccccccccccc@{}}
    \toprule
        \multicolumn{1}{l}{} & & \multicolumn{4}{c}{Cora-ML (LT)} & \multicolumn{4}{c}{Network Science (LT)} & \multicolumn{4}{c}{Power Grid (LT)} & \multicolumn{4}{c}{Jazz (LT)} \\
        \cmidrule(lr){3-6} \cmidrule(lr){7-10} \cmidrule(lr){11-14} \cmidrule(lr){15-18}
        \multicolumn{1}{c}{Training data} & Method & 1\% & 5\% & 10\% & 20\% & 1\% & 5\% & 10\% & 20\% & 1\% & 5\% & 10\% & 20\% & 1\% & 5\% & 10\% & 20\% \\
        \midrule
        1\% & \mymethod{} (U) & 18.1 & 80.8 & 93.4 & 98.8 & 6.9 & 26.1 & 42.4 & 66.8 & 7.6 & 30.9 & 58.0 & 88.4 & 2.0 & 23.0 & 58.7 & 100.0 \\
        \midrule
        5\% & \mymethod{} (U) & 19.0 & 80.7 & 92.8 & 98.9 & 6.9 & 25.9 & 41.8 & 66.2 & 7.6 & 30.6 & 58.3 & 88.3 & 2.0 & 22.7 & 56.8 & 100.0 \\
        \midrule
        20\% & \mymethod{} (U) & 19.3 & 80.7 & 93.0 & 98.7 & 6.8 & 26.3 & 42.3 & 67.0 & 7.6 & 30.7 & 58.2 & 88.1 & 2.0 & 30.4 & 54.5 & 100.0 \\
        \midrule
        100\% & \mymethod{} (U) & 18.9 & 80.0 & 93.3 & 98.8 & 6.8 & 26.5 & 41.7 & 66.4 & 7.6 & 31.0 & 58.4 & 88.1 & 2.0 & 30.4 & 48.0 & 100.0 \\
    \bottomrule
    \end{tabular}}

    \resizebox{\textwidth}{!}{
    \begin{tabular}{c|lcccccccccccccccc@{}}
    \toprule
        \multicolumn{1}{l}{} & & \multicolumn{4}{c}{Cora-ML (SIS)} & \multicolumn{4}{c}{Network Science (SIS)} & \multicolumn{4}{c}{Power Grid (SIS)} & \multicolumn{4}{c}{Jazz (SIS)} \\
        \cmidrule(lr){3-6} \cmidrule(lr){7-10} \cmidrule(lr){11-14} \cmidrule(lr){15-18}
        \multicolumn{1}{c}{Training data} & Method & 1\% & 5\% & 10\% & 20\% & 1\% & 5\% & 10\% & 20\% & 1\% & 5\% & 10\% & 20\% & 1\% & 5\% & 10\% & 20\% \\
        \midrule
        1\% & \mymethod{} (U) & 7.0 & 15.5 & 21.9 & 31.6 & 2.4 & 9.0 & 14.6 & 25.4 & 1.8 & 7.4 & 13.4 & 24.3 & 35.2 & 56.1 & 64.6 & 71.7 \\
        \midrule
        5\% & \mymethod{} (U) & 7.0 & 15.4 & 22.5 & 32.3 & 2.3 & 9.1 & 15.1 & 25.9 & 1.8 & 7.5 & 13.5 & 24.7 & 32.8 & 56.6 & 64.7 & 71.4 \\
        \midrule
        20\% & \mymethod{} (U) & 7.1 & 15.8 & 22.4 & 32.4 & 2.4 & 8.8 & 15.2 & 26.0 & 1.8 & 7.6 & 13.8 & 25.0 & 28.9 & 56.6 & 64.9 & 71.2 \\
        \midrule
        100\% & \mymethod{} (U) & 7.2 & 15.7 & 22.5 & 32.5 & 2.4 & 8.8 & 15.3 & 26.5 & 1.9 & 7.8 & 14.0 & 25.1 & 30.8 & 56.7 & 64.7 & 71.6 \\
    \bottomrule
    \end{tabular}}
\end{table}

\subsection{Data efficiency}\label{sec:data_efficiency}

Obtaining diffusion observations is another dominant practical bottleneck in learning-based IM. We therefore investigate how the performance of \mymethod{} varies as the amount of available training data decreases. Table~\ref{tab:data_efficiency} reports the influence spread obtained when training the surrogate using different fractions of the available observation pairs. The full set of results is provided in Appendix~\ref{app:data_efficiency_full}. Surprisingly, we observe that performance saturates rapidly, with only a small fraction of the training observations required to recover nearly the same influence spread as models trained on the complete dataset.\looseness=-1

This result has two important implications. First, the proposed surrogate exhibits strong sample efficiency, substantially reducing the cost of collecting diffusion observations. Second, the resulting reduction in surrogate training time enables a larger fraction of the computational budget to be allocated toward discrete search. On very large graphs, this redistribution of computation leads to significant reductions in total runtime (see Figure~\ref{fig:runtime_scalability}) while maintaining competitive IM performance.

\subsection{Mechanistic analysis of uniformly anchored node embeddings}

The previous experiments establish that \mymethod{} performs well empirically. We now investigate why such a lightweight surrogate generalizes effectively to previously unseen seed configurations. We also showcase the architectural robustness of \mymethod{} by evaluating an extremely lightweight variant where each node is represented using only its binary seed indicator in Appendix~\ref{app:binary_vs_8}.

\begin{table}[t]
    \centering
    \caption{Influence spread (\%) achieved by \mymethod{} with different node embedding initialization strategies and post-training embedding update rules.}
    \vspace{-3.5mm}
    \label{tab:init}
    \resizebox{\textwidth}{!}{
    \begin{tabular}{l|lcccccccccccccccc@{}}
    \toprule
        \multicolumn{1}{l}{} & & \multicolumn{4}{c}{Cora-ML (IC)} & \multicolumn{4}{c}{Network Science (IC)} & \multicolumn{4}{c}{Power Grid (IC)} & \multicolumn{4}{c}{Jazz (IC)} \\
        \cmidrule(lr){3-6} \cmidrule(lr){7-10} \cmidrule(lr){11-14} \cmidrule(lr){15-18}
        \multicolumn{1}{l}{Setting} & Method & 1\% & 5\% & 10\% & 20\% & 1\% & 5\% & 10\% & 20\% & 1\% & 5\% & 10\% & 20\% & 1\% & 5\% & 10\% & 20\% \\
        \midrule
        Random, no update 
        & \mymethod{} (U) & 16.4 & 32.4 & 42.4 & 56.3 & 9.4 & 23.7 & 36.5 & 56.9 & 7.9 & 25.7 & 39.8 & 58.7 & 10.4 & 25.3 & 36.1 & 50.5 \\
        \midrule
        Random, weighted avg 
        & \mymethod{} (U) & 14.8 & 32.4 & 38.9 & 53.0 & 9.2 & 23.7 & 36.0 & 49.6 & 7.9 & 25.4 & 38.5 & 55.0 & 10.6 & 26.8 & 36.3 & 49.4 \\
        \midrule
        Random, most similar
        & \mymethod{} (U) & 16.7 & 32.5 & 42.7 & 56.2 & 9.3 & \textbf{24.0} & \textbf{36.6} & 56.7 & 7.8 & 25.5 & 39.8 & 58.8 & 9.7 & 26.8 & 36.6 & 50.5 \\
        \midrule
        Random, global mean
        & \mymethod{} (U) & 14.4 & 32.3 & 38.6 & 53.3 & 9.3 & 23.7 & 36.1 & 49.4 & 8.0 & 25.3 & 38.8 & 54.7 & 10.4 & 26.5 & 35.9 & 49.4 \\
        \midrule
        Anchored, no update
        & \mymethod{} (U) & \textbf{16.9} & \textbf{33.3} & \textbf{43.1} & \textbf{57.3} & \textbf{9.7} & \textbf{24.0} & \textbf{36.6} & \textbf{57.8} & \textbf{8.2} & \textbf{26.2} & \textbf{40.4} & \textbf{59.5} & \textbf{12.3} & \textbf{27.6} & \textbf{37.5} & \textbf{51.1} \\
    \bottomrule
    \end{tabular}}
    
    \resizebox{\textwidth}{!}{
    \begin{tabular}{l|lcccccccccccccccc@{}}
    \toprule
        \multicolumn{1}{l}{} & & \multicolumn{4}{c}{Cora-ML (LT)} & \multicolumn{4}{c}{Network Science (LT)} & \multicolumn{4}{c}{Power Grid (LT)} & \multicolumn{4}{c}{Jazz (LT)} \\
        \cmidrule(lr){3-6} \cmidrule(lr){7-10} \cmidrule(lr){11-14} \cmidrule(lr){15-18}
        \multicolumn{1}{l}{Setting} & Method & 1\% & 5\% & 10\% & 20\% & 1\% & 5\% & 10\% & 20\% & 1\% & 5\% & 10\% & 20\% & 1\% & 5\% & 10\% & 20\% \\
        \midrule
        Random, no update 
        & \mymethod{} (U) & 15.3 & 79.1 & 90.9 & 97.6 & 5.6 & 22.0 & 35.2 & 59.4 & 6.6 & 26.7 & 52.7 & 83.1 & 1.8 & 7.7 & 24.0 & 99.5 \\
        \midrule
        Random, weighted avg 
        & \mymethod{} (U) & 16.9 & 74.1 & 84.0 & 89.3 & 4.7 & 21.0 & 29.8 & 43.0 & 7.0 & 29.2 & 51.3 & 75.8 & 1.9 & 7.1 & \textbf{68.6} & 94.9 \\
        \midrule
        Random, most similar
        & \mymethod{} (U) & 15.3 & 78.0 & 90.9 & 97.4 & 6.6 & 22.0 & 37.6 & 61.7 & 7.0 & 27.3 & 54.0 & 85.0 & \textbf{2.0} & 13.1 & 24.5 & 98.8 \\
        \midrule
        Random, global mean
        & \mymethod{} (U) & 16.1 & 73.7 & 84.4 & 89.6 & 4.9 & 21.1 & 28.9 & 42.3 & 7.1 & 29.8 & 51.1 & 75.6 & 1.9 & 7.1 & 66.4 & 99.5 \\
        \midrule
        Anchored, no update
        & \mymethod{} (U) & \textbf{18.9} & \textbf{80.0} & \textbf{93.3} & \textbf{98.8} & \textbf{6.8} & \textbf{26.5} & \textbf{41.7} & \textbf{66.4} & \textbf{7.6} & \textbf{31.0} & \textbf{58.4} & \textbf{88.1} & \textbf{2.0} & \textbf{30.4} & 48.0 & \textbf{100.0} \\
    \bottomrule
    \end{tabular}}
    
    \resizebox{\textwidth}{!}{
    \begin{tabular}{l|lcccccccccccccccc@{}}
    \toprule
        \multicolumn{1}{l}{} & & \multicolumn{4}{c}{Cora-ML (SIS)} & \multicolumn{4}{c}{Network Science (SIS)} & \multicolumn{4}{c}{Power Grid (SIS)} & \multicolumn{4}{c}{Jazz (SIS)} \\
        \cmidrule(lr){3-6} \cmidrule(lr){7-10} \cmidrule(lr){11-14} \cmidrule(lr){15-18}
        \multicolumn{1}{l}{Setting} & Method & 1\% & 5\% & 10\% & 20\% & 1\% & 5\% & 10\% & 20\% & 1\% & 5\% & 10\% & 20\% & 1\% & 5\% & 10\% & 20\% \\
        \midrule
        Random, no update 
        & \mymethod{} (U) & 6.8 & 15.5 & 22.5 & \textbf{32.9} & 2.1 & 8.1 & 14.5 & 25.8 & 1.6 & 7.2 & 13.3 & 24.4 & 31.6 & 54.6 & 62.8 & 70.8 \\
        \midrule
        Random, weighted avg 
        & \mymethod{} (U) & \textbf{7.2} & 15.8 & 21.9 & 31.2 & \textbf{2.9} & 8.9 & 14.4 & 23.6 & 1.8 & 7.7 & 13.8 & 24.2 & 32.3 & 54.0 & 61.6 & 69.9 \\
        \midrule
        Random, most similar
        & \mymethod{} (U) & 7.1 & 15.7 & \textbf{22.6} & 32.6 & 2.7 & \textbf{9.0} & 15.3 & 26.2 & 1.8 & 7.4 & 13.6 & 24.4 & 31.1 & 53.2 & 62.7 & 70.0 \\
        \midrule
        Random, global mean
        & \mymethod{} (U) & \textbf{7.2} & \textbf{15.8} & 21.8 & 31.2 & \textbf{2.9} & 8.8 & 14.5 & 23.6 & 1.8 & 7.7 & 13.8 & 24.2 & \textbf{32.7} & 54.2 & 61.5 & 68.5 \\
        \midrule
        Anchored, no update
        & \mymethod{} (U) & \textbf{7.2} & 15.7 & 22.5 & 32.5 & 2.4 & 8.8 & \textbf{15.3} & \textbf{26.5} & \textbf{1.9} & \textbf{7.8} & \textbf{14.0} & \textbf{25.1} & 30.8 & \textbf{56.7} & \textbf{64.7} & \textbf{71.6} \\
    \bottomrule
    \end{tabular}}
\end{table}

\textbf{Effect of initialization strategy on seed set quality.}
The proposed framework initializes every node embedding from a common random anchor before learning topology and diffusion-dependent representations. To evaluate the importance of this design, we compare against conventional independent random initialization, where each node begins from a distinct embedding. We additionally consider three post-training update strategies for handling unseen nodes: 1) computing a graph structural feature similarity weighted average over the embeddings of seen nodes, 2) assigning the embedding of the single most structurally similar seen node, and 3) assigning the global mean of the embeddings of seen nodes. For the graph structural features, we include degree centrality, PageRank~\citep{page1998pagerank}, clustering coefficient~\citep{watts1998collective}, betweenness centrality~\cite{freeman1977betweenness}, and core number~\citep{batagelj2003corenum}. Weights and similarities are computed using cosine similarity.\looseness=-1

Table~\ref{tab:init} shows that uniformly anchored initialization consistently produces higher downstream influence spread. The full set of results is provided in Appendix~\ref{app:init_full}. While unseen node update strategies may improve the performance of independently initialized embeddings, the uniformly anchored approach achieves the strongest overall performance without requiring any post hoc operation. Furthermore, as revealed in Table~\ref{tab:update_rules} in Appendix~\ref{app:update_rules}, when employing uniformly anchored node embeddings, there is no clear benefit of utilizing custom update rules for the unseen nodes; this design already robustly handles them. 

\textbf{Geometric interpretation of the embedding space.}
To further understand the learned representations, we visualize the embedding space produced by surrogate training under different initialization strategies. For direct interpretability, we set the node embedding dimension to $d = 2$, allowing us to plot the learned latent space directly without relying on dimensionality reduction. We confirm in Table~\ref{tab:emb_dim_2_vs_8} in Appendix~\ref{app:emb_dim_2_vs_8} that the $d = 2$ variant maintains high downstream IM performance.

\begin{figure}[!t]
    \centering
    \begin{subfigure}{0.32\textwidth}
        \centering
        \includegraphics[width=\linewidth]{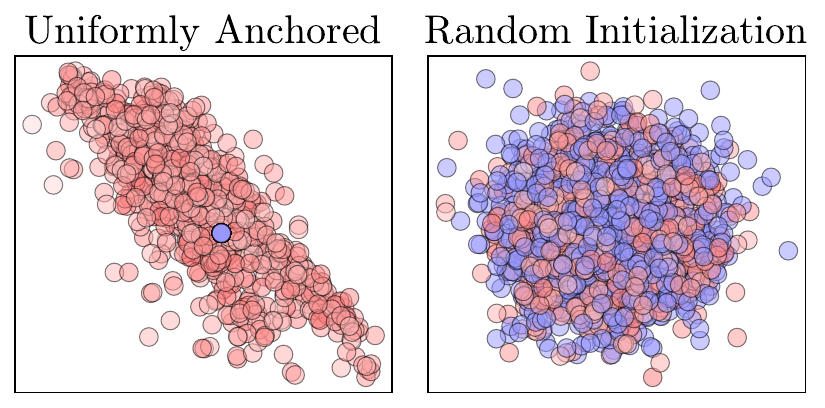}
        \caption{IC}
    \end{subfigure}
    \hfill
    \begin{subfigure}{0.32\textwidth}
        \centering
        \includegraphics[width=\linewidth]{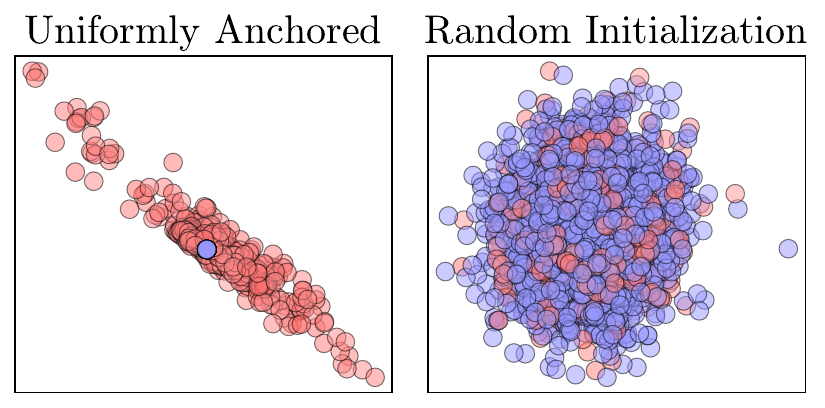}
        \caption{LT}
    \end{subfigure}
    \hfill
    \begin{subfigure}{0.32\textwidth}
        \centering
        \includegraphics[width=\linewidth]{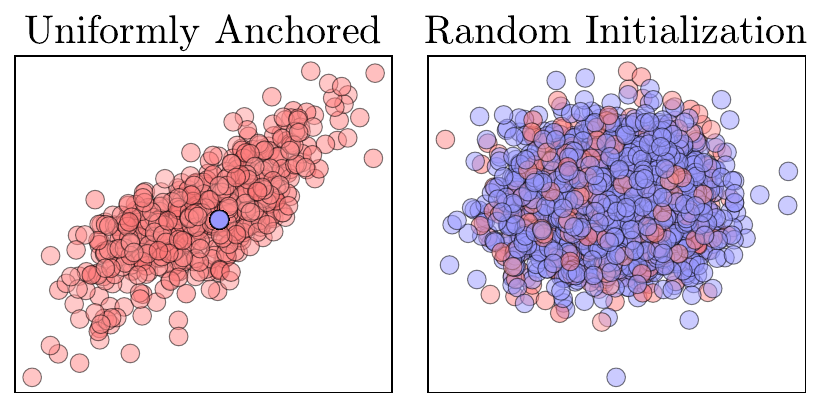}
        \caption{SIS}
    \end{subfigure}
    \vspace{-3mm}
    \caption{Learned node embeddings of Cora-ML. Red: seen nodes; blue: unseen nodes.}
    \label{fig:emb_coraml}
\end{figure}

\begin{figure}[!t]
    \centering
    \begin{subfigure}{0.32\textwidth}
        \centering
        \includegraphics[width=\linewidth]{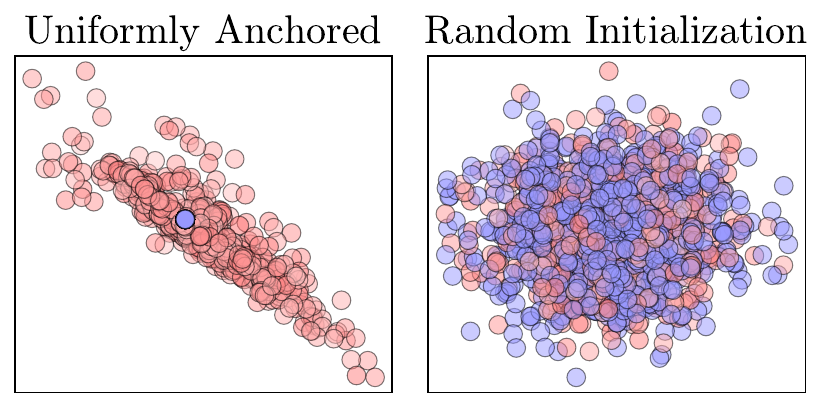}
        \caption{IC}
    \end{subfigure}
    \hfill
    \begin{subfigure}{0.32\textwidth}
        \centering
        \includegraphics[width=\linewidth]{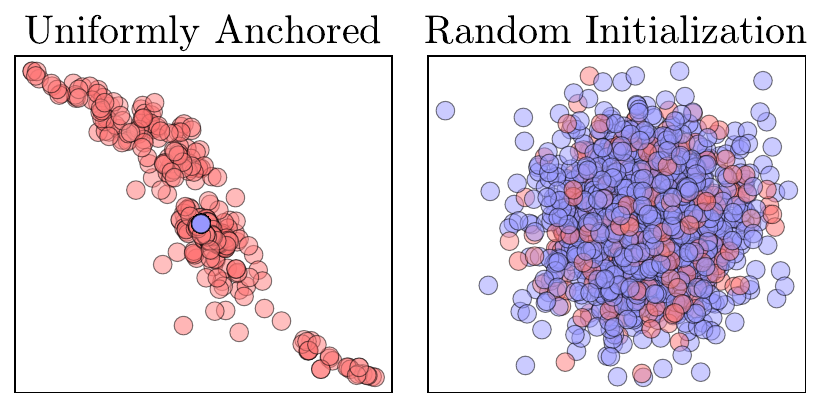}
        \caption{LT}
    \end{subfigure}
    \hfill
    \begin{subfigure}{0.32\textwidth}
        \centering
        \includegraphics[width=\linewidth]{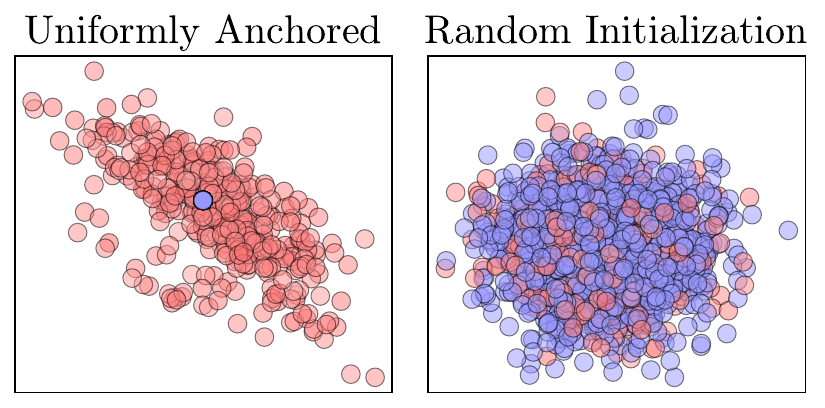}
        \caption{SIS}
    \end{subfigure}
    \vspace{-3mm}
    \caption{Learned node embeddings of Network Science. Red: seen nodes; blue: unseen nodes.}
    \label{fig:emb_netscience}
\end{figure}

\begin{figure}[!t]
    \centering
    \begin{subfigure}{0.32\textwidth}
        \centering
        \includegraphics[width=\linewidth]{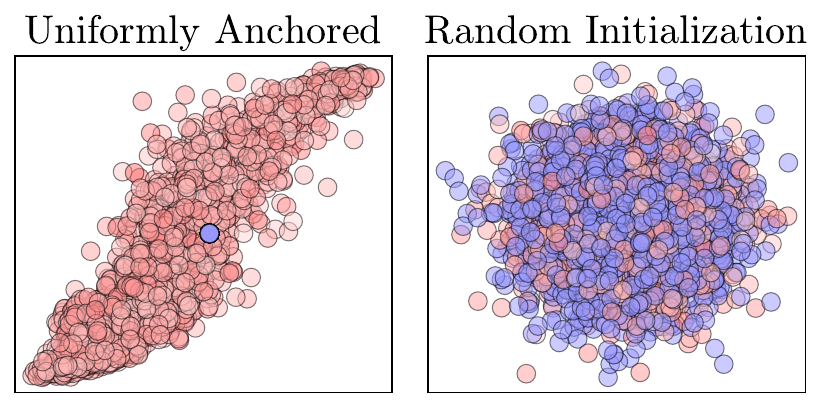}
        \caption{IC}
    \end{subfigure}
    \hfill
    \begin{subfigure}{0.32\textwidth}
        \centering
        \includegraphics[width=\linewidth]{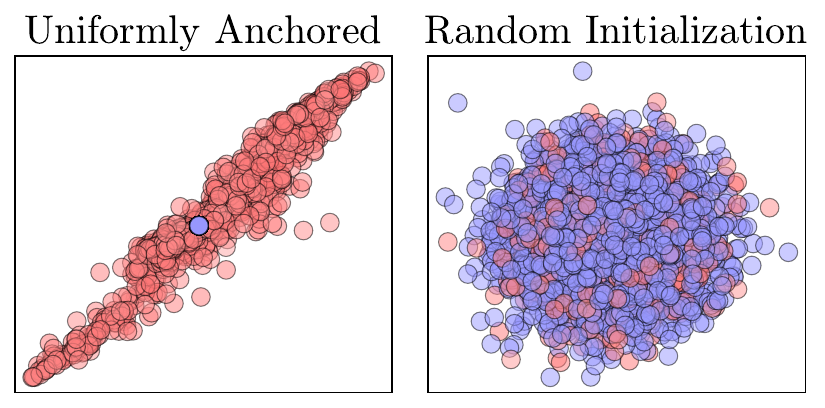}
        \caption{LT}
    \end{subfigure}
    \hfill
    \begin{subfigure}{0.32\textwidth}
        \centering
        \includegraphics[width=\linewidth]{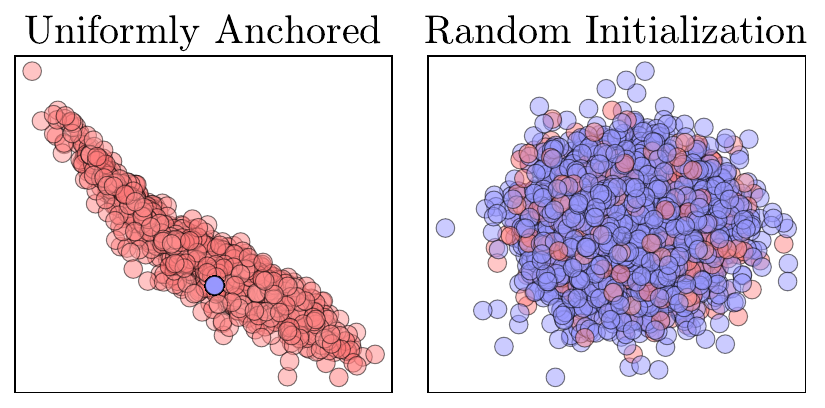}
        \caption{SIS}
    \end{subfigure}
    \vspace{-3mm}
    \caption{Learned node embeddings of Power Grid. Red: seen nodes; blue: unseen nodes.}
    \label{fig:emb_powergrid}
\end{figure}

\begin{figure}[!t]
    \centering
    \begin{subfigure}{0.32\textwidth}
        \centering
        \includegraphics[width=\linewidth]{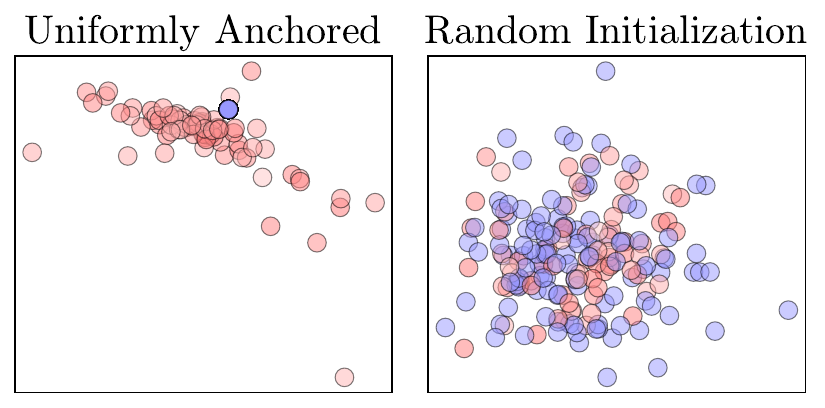}
        \caption{IC}
    \end{subfigure}
    \hfill
    \begin{subfigure}{0.32\textwidth}
        \centering
        \includegraphics[width=\linewidth]{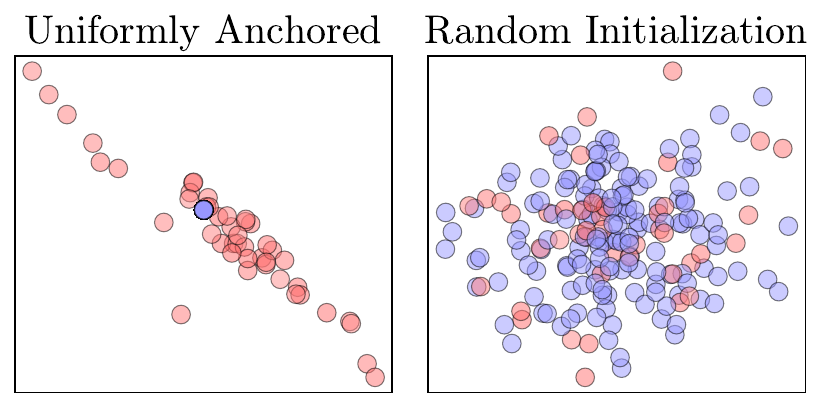}
        \caption{LT}
    \end{subfigure}
    \hfill
    \begin{subfigure}{0.32\textwidth}
        \centering
        \includegraphics[width=\linewidth]{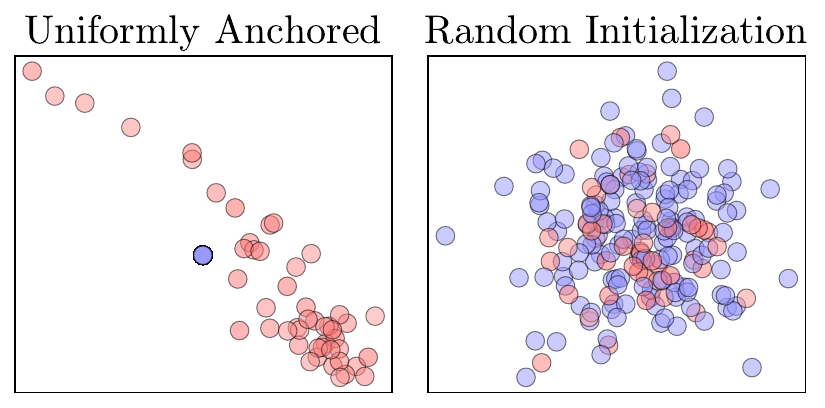}
        \caption{SIS}
    \end{subfigure}
    \vspace{-3mm}
    \caption{Learned node embeddings of Jazz. Red: seen nodes; blue: unseen nodes.}
    \label{fig:emb_jazz}
\end{figure}

As shown in Figures~\ref{fig:emb_coraml}, \ref{fig:emb_netscience}, \ref{fig:emb_powergrid}, and \ref{fig:emb_jazz}, uniformly anchored node embeddings often naturally evolve into a smooth, highly structured low-dimensional manifold after training, forming an elliptical distribution, with the embeddings of the unseen nodes centered. Because every node starts from the same initialization anchor, learning is driven purely by graph topology and diffusion pattern. 
In contrast, independently initialized embeddings form an isotropic distribution whose higher variance results in a less structured embedding distribution, with the embeddings of the seen and unseen nodes scattered.\looseness=-1

We identify two advantages of the resulting embedding space geometry: 1) The low-dimensional manifold supports linear scalability. When the GNN aggregates neighborhoods, the embeddings naturally scale linearly, giving a strong and explicit signal of local influence density; 2) Acting as a special token, the anchor forces the embeddings of the unseen nodes to have the same values, which encourages the active neighbors of such nodes to dictate their role in influence spread prediction and makes it easier for the GNN to meaningfully override their embeddings through message passing. Together, these geometric properties induced by the uniform anchor prevent downstream discrete search from encountering uncalibrated local maxima, explaining why a lightweight GNN surrogate can support robust combinatorial optimization in discrete space.

\section{Conclusion}

In this work, we presented \mymethod{}, a paradigm for IM that shifts computational effort from training complex model architectures to direct discrete search. By pairing uniformly anchored node embeddings with a shallow two-layer GNN surrogate, \mymethod{} robustly predicts complex infection dynamics in a single forward pass without requiring explicit parameterization of propagation rules. Combined with surrogate-guided batched multi-swap simulated annealing, the framework directly explores the combinatorial seed space, completely bypassing gradient computations and continuous relaxations. Empirical results demonstrate that \mymethod{} drastically cuts overall time-to-solution while consistently achieving superior influence spread and remarkable data efficiency. 

\newpage

\bibliography{bibliography}

@inproceedings{ling2023deepim,
  title={Deep graph representation learning and optimization for influence maximization},
  author={Ling, Chen and Jiang, Junji and Wang, Junxiang and Thai, My T and Xue, Renhao and Song, James and Qiu, Meikang and Zhao, Liang},
  booktitle={International conference on machine learning},
  pages={21350--21361},
  year={2023},
  organization={PMLR}
}

@article{hevapathige2025deepsn,
  title={Deepsn: a sheaf neural framework for influence maximization},
  author={Hevapathige, Asela and Wang, Qing and Zehmakan, Ahad N},
  journal={Proceedings of the AAAI Conference on Artificial Intelligence},
  volume={39},
  number={16},
  pages={17177--17185},
  year={2025}
}

@article{hamilton2017graphsage,
  title={Inductive representation learning on large graphs},
  author={Hamilton, Will and Ying, Zhitao and Leskovec, Jure},
  journal={Advances in neural information processing systems},
  volume={30},
  year={2017}
}

@inproceedings{tang2015imm,
  title={Influence maximization in near-linear time: A martingale approach},
  author={Tang, Youze and Shi, Yanchen and Xiao, Xiaokui},
  booktitle={Proceedings of the 2015 ACM SIGMOD international conference on management of data},
  pages={1539--1554},
  year={2015}
}

@inproceedings{
    velickovic2018graph,
    title={Graph Attention Networks},
    author={Petar Veličković and Guillem Cucurull and Arantxa Casanova and Adriana Romero and Pietro Liò and Yoshua Bengio},
    booktitle={International Conference on Learning Representations},
    year={2018}
}

@article{rossi2015network,
  title={The network data repository with interactive graph analytics and visualization},
  author={Rossi, Ryan and Ahmed, Nesreen},
  journal={Proceedings of the AAAI conference on artificial intelligence},
  volume={29},
  number={1},
  year={2015}
}

@article{mccallum2000automating,
  title={Automating the construction of internet portals with machine learning},
  author={McCallum, Andrew Kachites and Nigam, Kamal and Rennie, Jason and Seymore, Kristie},
  journal={Information Retrieval},
  volume={3},
  number={2},
  pages={127--163},
  year={2000},
  publisher={Springer}
}

@article{lerman2012social,
  title={Social contagion: An empirical study of information spread on digg and twitter follower graphs},
  author={Lerman, Kristina and Ghosh, Rumi and Surachawala, Tawan},
  journal={arXiv preprint arXiv:1202.3162},
  year={2012}
}

@article{kermack1927epidemics,
 ISSN = {09501207},
 author = {W. O. Kermack and A. G. McKendrick},
 journal = {Proceedings of the Royal Society of London. Series A, Containing Papers of a Mathematical and Physical Character},
 number = {772},
 pages = {700--721},
 publisher = {The Royal Society},
 title = {A Contribution to the Mathematical Theory of Epidemics},
 urldate = {2026-07-29},
 volume = {115},
 year = {1927}
}

@article{panagopoulos2020IMINFECTOR,
  title={Multi-task learning for influence estimation and maximization},
  author={Panagopoulos, George and Malliaros, Fragkiskos D and Vazirgiannis, Michalis},
  journal={IEEE Transactions on Knowledge and Data Engineering},
  volume={34},
  number={9},
  pages={4398--4409},
  year={2020},
  publisher={IEEE}
}

@article{li2022piano,
  title={PIANO: Influence maximization meets deep reinforcement learning},
  author={Li, Hui and Xu, Mengting and Bhowmick, Sourav S and Rayhan, Joty Shafiq and Sun, Changsheng and Cui, Jiangtao},
  journal={IEEE Transactions on Computational Social Systems},
  volume={10},
  number={3},
  pages={1288--1300},
  year={2022},
  publisher={IEEE}
}

@article{chen2023touplegdd,
  title={ToupleGDD: A fine-designed solution of influence maximization by deep reinforcement learning},
  author={Chen, Tiantian and Yan, Siwen and Guo, Jianxiong and Wu, Weili},
  journal={IEEE Transactions on Computational Social Systems},
  volume={11},
  number={2},
  pages={2210--2221},
  year={2023},
  publisher={IEEE}
}

@inproceedings{tang2018opim,
  title={Online processing algorithms for influence maximization},
  author={Tang, Jing and Tang, Xueyan and Xiao, Xiaokui and Yuan, Junsong},
  booktitle={Proceedings of the 2018 international conference on management of data},
  pages={991--1005},
  year={2018}
}

@inproceedings{guo2020subsim,
  title={Influence maximization revisited: Efficient reverse reachable set generation with bound tightened},
  author={Guo, Qintian and Wang, Sibo and Wei, Zhewei and Chen, Ming},
  booktitle={Proceedings of the 2020 ACM SIGMOD international conference on management of data},
  pages={2167--2181},
  year={2020}
}

@inproceedings{zhang2026dynaflux,
  title={DynaFLUX: Implicit Dynamics-Preserving Reinforcement Learning for Topology-Free Influence Maximization},
  author={Zhang, Daiyunke and Deng, Ting and Zhu, Tianchen and Ma, Shuai and Li, Daqing and Peng, Mingtian and Tian, Feng},
  booktitle={Proceedings of the ACM Web Conference 2026},
  pages={4940--4951},
  year={2026}
}

@inproceedings{lei2015oim,
  title={Online influence maximization},
  author={Lei, Siyu and Maniu, Silviu and Mo, Luyi and Cheng, Reynold and Senellart, Pierre},
  booktitle={Proceedings of the 21th ACM SIGKDD international conference on knowledge discovery and data mining},
  pages={645--654},
  year={2015}
}

@inproceedings{chen2010scalable,
  title={Scalable influence maximization for prevalent viral marketing in large-scale social networks},
  author={Chen, Wei and Wang, Chi and Wang, Yajun},
  booktitle={Proceedings of the 16th ACM SIGKDD international conference on Knowledge discovery and data mining},
  pages={1029--1038},
  year={2010}
}

@inproceedings{kempe2003maximizing,
  title={Maximizing the spread of influence through a social network},
  author={Kempe, David and Kleinberg, Jon and Tardos, {\'E}va},
  booktitle={Proceedings of the ninth ACM SIGKDD international conference on Knowledge discovery and data mining},
  pages={137--146},
  year={2003}
}

@article{neophytou2024promoting,
  title={Promoting fair vaccination strategies through influence maximization: a case study on COVID-19 spread},
  author={Neophytou, Nicola and Ta{\"\i}k, Afaf and Farnadi, Golnoosh},
  journal={Proceedings of the AAAI Conference on Artificial Intelligence},
  volume={38},
  number={20},
  pages={22285--22293},
  year={2024}
}

@article{granovetter1978threshold,
 ISSN = {00029602, 15375390},
 author = {Mark Granovetter},
 journal = {American Journal of Sociology},
 number = {6},
 pages = {1420--1443},
 publisher = {University of Chicago Press},
 title = {Threshold Models of Collective Behavior},
 urldate = {2026-08-02},
 volume = {83},
 year = {1978}
}

@article{batagelj2003corenum,
  title={An o (m) algorithm for cores decomposition of networks},
  author={Batagelj, Vladimir and Zaversnik, Matjaz},
  journal={arXiv preprint cs/0310049},
  year={2003}
}

@article{saito2012efficient,
  title={Efficient discovery of influential nodes for SIS models in social networks},
  author={Saito, Kazumi and Kimura, Masahiro and Ohara, Kouzou and Motoda, Hiroshi},
  journal={Knowledge and information systems},
  volume={30},
  number={3},
  pages={613--635},
  year={2012},
  publisher={Springer}
}

@inproceedings{tang2014influence,
  title={Influence maximization: Near-optimal time complexity meets practical efficiency},
  author={Tang, Youze and Xiao, Xiaokui and Shi, Yanchen},
  booktitle={Proceedings of the 2014 ACM SIGMOD international conference on Management of data},
  pages={75--86},
  year={2014}
}

@inproceedings{nguyen2016stop,
  title={Stop-and-stare: Optimal sampling algorithms for viral marketing in billion-scale networks},
  author={Nguyen, Hung T and Thai, My T and Dinh, Thang N},
  booktitle={Proceedings of the 2016 international conference on management of data},
  pages={695--710},
  year={2016}
}

@inproceedings{borgs2014maximizing,
  title={Maximizing social influence in nearly optimal time},
  author={Borgs, Christian and Brautbar, Michael and Chayes, Jennifer and Lucier, Brendan},
  booktitle={Proceedings of the twenty-fifth annual ACM-SIAM symposium on Discrete algorithms},
  pages={946--957},
  year={2014},
  organization={SIAM}
}

@article{li2019tiptop,
  title={Tiptop:(almost) exact solutions for influence maximization in billion-scale networks},
  author={Li, Xiang and Smith, J David and Dinh, Thang N and Thai, My T},
  journal={IEEE/ACM Transactions on Networking},
  volume={27},
  number={2},
  pages={649--661},
  year={2019},
  publisher={IEEE}
}

@inproceedings{leskovec2007cost,
  title={Cost-effective outbreak detection in networks},
  author={Leskovec, Jure and Krause, Andreas and Guestrin, Carlos and Faloutsos, Christos and VanBriesen, Jeanne and Glance, Natalie},
  booktitle={Proceedings of the 13th ACM SIGKDD international conference on Knowledge discovery and data mining},
  pages={420--429},
  year={2007}
}

@article{hong2020efficient,
  title={Efficient minimum cost seed selection with theoretical guarantees for competitive influence maximization},
  author={Hong, Wenjing and Qian, Chao and Tang, Ke},
  journal={IEEE transactions on cybernetics},
  volume={51},
  number={12},
  pages={6091--6104},
  year={2020},
  publisher={IEEE}
}

@inproceedings{lin2015learning,
  title={A learning-based framework to handle multi-round multi-party influence maximization on social networks},
  author={Lin, Su-Chen and Lin, Shou-De and Chen, Ming-Syan},
  booktitle={Proceedings of the 21th ACM SIGKDD international conference on knowledge discovery and data mining},
  pages={695--704},
  year={2015}
}

@article{li2019disco,
  title={Disco: Influence maximization meets network embedding and deep learning},
  author={Li, Hui and Xu, Mengting and Bhowmick, Sourav S and Sun, Changsheng and Jiang, Zhongyuan and Cui, Jiangtao},
  journal={arXiv preprint arXiv:1906.07378},
  year={2019}
}

@article{tian2020deep,
  title={Deep Reinforcement Learning-Based Approach to Tackle Topic-Aware Influence Maximization},
  author={Tian, Shan and Mo, Songsong and Wang, Liwei and Peng, Zhiyong},
  journal={Data Science and Engineering},
  volume={5},
  number={1},
  pages={1--11},
  year={2020},
  publisher={Springer}
}

@article{manchanda2020gcomb,
  title={Gcomb: Learning budget-constrained combinatorial algorithms over billion-sized graphs},
  author={Manchanda, Sahil and Mittal, Akash and Dhawan, Anuj and Medya, Sourav and Ranu, Sayan and Singh, Ambuj},
  journal={Advances in Neural Information Processing Systems},
  volume={33},
  pages={20000--20011},
  year={2020}
}

@article{kumar2022influence,
  title={Influence maximization in social networks using graph embedding and graph neural network},
  author={Kumar, Sanjay and Mallik, Abhishek and Khetarpal, Anavi and Panda, Bhawani Sankar},
  journal={Information Sciences},
  volume={607},
  pages={1617--1636},
  year={2022},
  publisher={Elsevier}
}

@inproceedings{zhang2026imgnn,
  title={IMGNN: An Efficient, Effective and Generalizable Algorithm for Influence Maximization in Social Networks},
  author={Zhang, Haotian and Han, Kai and Yin, Zhizhuo and Cui, Shuang and Tang, Jing and Hui, Pan},
  booktitle={Proceedings of the 32nd ACM SIGKDD Conference on Knowledge Discovery and Data Mining V. 1},
  pages={1904--1915},
  year={2026}
}

@inproceedings{page1998pagerank,
  title={The pagerank citation ranking: Bring order to the web},
  author={Page, Lawrence and Brin, Sergey and Motwani, Rajeev and Winograd, Terry},
  booktitle={Proc. of the 7th International World Wide Web Conf},
  year={1998}
}

@article{freeman1977betweenness,
  title={A set of measures of centrality based on betweenness},
  author={Freeman, Linton C},
  journal={Sociometry},
  pages={35--41},
  year={1977},
  publisher={JSTOR}
}

@article{watts1998collective,
  title={Collective dynamics of ‘small-world’networks},
  author={Watts, Duncan J and Strogatz, Steven H},
  journal={nature},
  volume={393},
  number={6684},
  pages={440--442},
  year={1998},
  publisher={Nature Publishing Group UK London}
}

@inproceedings{adam,
  author       = {Diederik P. Kingma and Jimmy Ba},
  title        = {Adam: {A} Method for Stochastic Optimization},
  booktitle    = {International Conference on Learning Representations},
  year         = {2015},
}

@inproceedings{loshchilov2018adamw,
    title={Decoupled Weight Decay Regularization},
    author={Ilya Loshchilov and Frank Hutter},
    booktitle={International Conference on Learning Representations},
    year={2019},
}

@inproceedings{fey2019pyg,
  title={Fast Graph Representation Learning with {PyTorch Geometric}},
  author={Fey, Matthias and Lenssen, Jan E.},
  booktitle={ICLR Workshop on Representation Learning on Graphs and Manifolds},
  year={2019},
}

@inproceedings{fey2025pyg2,
    title={PyG 2.0: Scalable Learning on Real World Graphs},
    author={Matthias Fey and Jinu Sunil and Akihiro Nitta and Rishi Puri and Manan Shah and Bla{\v{z}} Stojanovi{\v{c}} and Ramona Bendias and Alexandria Barghi and Vid Kocijan and Zecheng Zhang and Xinwei He and Jan Eric Lenssen and Jure Leskovec},
    booktitle={Temporal Graph Learning Workshop @ KDD 2025},
    year={2025},
}

@inproceedings{pytorch,
    author = {Paszke, Adam and Gross, Sam and Massa, Francisco and Lerer, Adam and Bradbury, James and Chanan, Gregory and Killeen, Trevor and Lin, Zeming and Gimelshein, Natalia and Antiga, Luca and Desmaison, Alban and K\"{o}pf, Andreas and Yang, Edward and DeVito, Zach and Raison, Martin and Tejani, Alykhan and Chilamkurthy, Sasank and Steiner, Benoit and Fang, Lu and Bai, Junjie and Chintala, Soumith},
    title = {PyTorch: an imperative style, high-performance deep learning library},
    year = {2019},
    booktitle = {Proceedings of the 33rd International Conference on Neural Information Processing Systems},
}

@inproceedings{he2016deep,
  title={Deep residual learning for image recognition},
  author={He, Kaiming and Zhang, Xiangyu and Ren, Shaoqing and Sun, Jian},
  booktitle={Proceedings of the IEEE conference on computer vision and pattern recognition},
  pages={770--778},
  year={2016}
}

@inproceedings{hansen2020sheaf,
    title={Sheaf Neural Networks},
    author={Jakob Hansen and Thomas Gebhart},
    booktitle={TDA {\&} Beyond},
    year={2020},
}

@article{bodnar2022sheaf,
  title={Neural sheaf diffusion: A topological perspective on heterophily and oversmoothing in gnns},
  author={Bodnar, Cristian and Di Giovanni, Francesco and Chamberlain, Benjamin and Lio, Pietro and Bronstein, Michael},
  journal={Advances in Neural Information Processing Systems},
  volume={35},
  pages={18527--18541},
  year={2022}
}
\bibliographystyle{iclr2027_conference}

\appendix

\newpage
\appendix

\section{Additional literature review}\label{app:related}
\textbf{DeepIM.}
DeepIM~\citep{ling2023deepim} formulates IM as a continuous optimization problem. It employs an autoencoder to map discrete seed vectors into a continuous latent representation. In parallel, it trains a graph attention network~\citep{velickovic2018graph} to predict diffusion outcomes from binary seed configurations. During inference, it optimizes seed probabilities in the latent space through gradient updates to maximize the predicted final spread, projects the latent representation back to a seed vector, and applies a top-$k$ projection to recover a discrete seed set. Although this approach enables efficient inference, the continuous latent seed set representations introduce additional learning overhead and a gap between continuous latent variables and discrete seed selections.

\textbf{DeepSN.}
DeepSN~\citep{hevapathige2025deepsn} follows a similar surrogate-based optimization framework but focuses on improving diffusion prediction accuracy through a more sophisticated graph architecture. Specifically, it replaces conventional message-passing networks with a sheaf neural network~\citep{hansen2020sheaf, bodnar2022sheaf} that takes in binary seed set vectors with one-hot-coded unique node degrees and incorporates an additional learnable node scoring mechanism during inference. While these architectural improvements enhance predictive capacity, they also introduce training complexity.

\textbf{Limitations of existing approaches.}
Although existing neural surrogate frameworks demonstrate the feasibility of diffusion-model-agnostic IM, they rely on two implicit assumptions. First, they assume that high-quality seed set optimization requires continuous relaxation of the discrete search space. Second, they assume that increasingly complex neural architectures are necessary to accurately model diffusion dynamics.

In contrast, \mymethod{} investigates whether a lightweight neural surrogate combined with direct discrete optimization can achieve competitive or superior performance. Rather than optimizing a continuous representation of seed sets, \mymethod{} operates directly in the original combinatorial space using batched multi-swap simulated annealing guided by the learned surrogate. Our design removes latent seed representations and specialized architectures while enabling efficient and effective exploration of large-scale seed spaces.

\section{Datasets}\label{app:datasets}

\begin{table}[h]
    \centering
    \caption{Dataset statistics. The number of message-passing edges is reported.}
    \label{tab:data}
    \resizebox{0.7\textwidth}{!}{
    \begin{tabular}{lllllll}
        \toprule
        & Jazz & NS & Cora-ML & PG & Synthetic & Digg \\
        \midrule
        Nodes & 198  & 1,589 & 2,810  & 4941  & 50,000  & 279,630 \\
        Edges & 5,484 & 5,484 & 15,962 & 13188 & 500,000 & 3,096,252 \\
        \bottomrule
    \end{tabular}}
\end{table}

For diffusion observation generation and IM performance evaluation, we run simulations for 100 diffusion steps on the graphs with each of the diffusion models and record the final node states as the diffusion outcomes. For the LT model, the thresholds are sampled uniformly from $[0.3, 0.6]$; for the IC model, the propagation probabilities are set to be $p_{u,v} = 1/d_v^{\mathrm{in}}, \forall (u, v) \in E$, where $d_v^{\mathrm{in}}$ is the in-degree of node $v$; for the SIS model, the infection and recovery probabilities are both set to be 0.001. The diffusion configurations and simulation settings are consistent with prior works~\citep{ling2023deepim, hevapathige2025deepsn}.

\section{A note on final spread computation for evaluation}\label{app:eval}

The final spread is obtained by running simulations 100 times with the seed sets produced by the methods and averaging the results, which we view as one round of simulations. However, there can still be noticeable variance across rounds based on our observation. To further reduce the effect of randomness, we perform 5 rounds of such simulations (i.e., 5 $\times$ 100) and report the average over the 5 rounds (i.e., the average of the average spread). The results showing standard deviations are provided in Appendix~\ref{app:std}.

\section{Implementation details}\label{app:impl_details}

\textbf{Uniformly anchored node embeddings and lightweight neural surrogate.}
Unless otherwise specified, the node embedding dimension is fixed at $d = 8$. The neural surrogate consists of two GraphSAGE layers~\citep{hamilton2017graphsage} followed by a linear prediction head. Both GraphSAGE layers use a hidden dimension of 128, and a residual connection~\citep{he2016deep} is added to the second layer. All models are trained for 200 epochs using AdamW~\citep{adam, loshchilov2018adamw} with a learning rate of $5e^{-4}$ and a weight decay of $1e^{-4}$. The same model architecture and training hyperparameters are used throughout all experiments.

\textbf{Batched multi-swap simulated annealing.}
The search hyperparameters are selected according to the graph size and search complexity. For Jazz, Cora-ML, Network Science, and Power Grid, we set the number of steps to $I = 10,000$, the batch size to $B = 4$, and the number of swaps $r = 1$. For Synthetic, we increase the search budget to $I = 50{,}000$ and decrease the batch size to $B = 1$ while keeping $r = 1$. For Digg, we use $I = 20{,}000$, $B = 1$, and $r = 20$ to enable more aggressive exploration of the substantially larger search space.

All models are implemented using PyTorch~\citep{pytorch} and PyTorch Geometric~\citep{fey2019pyg, fey2025pyg2} libraries. Experiments are conducted on a machine with a single NVIDIA GeForce RTX 4090 GPU, a 32-core Intel Core i9-14900K CPU, and 64 GB of RAM running Ubuntu 24.04.

\section{Additional experimental results}

\subsection{Influence maximization performance of \mymethod{} with different temperature schedules}\label{app:temperature}

\begin{table}[h]
    \centering
    \caption{Influence spread (\%) achieved by \mymethod{} with different temperature schedules.}
    \label{tab:temperature}
    \resizebox{\textwidth}{!}{
    \begin{tabular}{l|lcccccccccccccccc@{}}
    \toprule
        \multicolumn{1}{l}{} & & \multicolumn{4}{c}{Cora-ML (IC)} & \multicolumn{4}{c}{Network Science (IC)} & \multicolumn{4}{c}{Power Grid (IC)} & \multicolumn{4}{c}{Jazz (IC)} \\
        \cmidrule(lr){3-6} \cmidrule(lr){7-10} \cmidrule(lr){11-14} \cmidrule(lr){15-18}
        \multicolumn{1}{l}{Setting} & Method & 1\% & 5\% & 10\% & 20\% & 1\% & 5\% & 10\% & 20\% & 1\% & 5\% & 10\% & 20\% & 1\% & 5\% & 10\% & 20\% \\
        \midrule
        \multirow{3}{*}{\shortstack[l]{$T_0 = 100$\\$\alpha = 0.2$}}
        & \mymethod{} (C) & 14.2 & 33.2 & 40.0 & 54.5 & 9.4 & 23.8 & 36.7 & 50.3 & 8.4 & 26.6 & 40.6 & 57.8 & 9.8 & 27.1 & 36.5 & 50.2 \\
        & \mymethod{} (U) & 16.7 & 33.2 & 43.1 & 56.9 & \textbf{9.7} & 24.2 & 37.1 & 57.8 & 8.2 & 26.1 & 40.6 & 59.5 & 12.1 & \textbf{27.3} & \textbf{37.7} & 50.1 \\
        & \mymethod{} (U*) & 16.7 & \textbf{33.6} & 43.2 & 57.1 & 9.6 & 24.3 & \textbf{37.2} & \textbf{58.2} & 8.4 & 26.8 & \textbf{41.4} & \textbf{60.1} & 11.8 & \textbf{27.3} & 37.1 & 50.7 \\
        \midrule
        \multirow{3}{*}{\shortstack[l]{$T_0 = 1e^{-6}$\\$\alpha = 1$}}
        & \mymethod{} (C) & 14.2 & 33.2 & 39.8 & 54.7 & 9.5 & 24.0 & 36.6 & 50.3 & \textbf{8.5} & 26.7 & 40.5 & 57.6 & 10.4 & 27.2 & 36.7 & 50.0 \\
        & \mymethod{} (U) & 16.9 & 33.3 & 43.1 & 57.3 & \textbf{9.7} & 24.0 & 36.6 & 57.8 & 8.2 & 26.2 & 40.4 & 59.5 & \textbf{12.3} & 27.6 & 37.5 & \textbf{51.1} \\
        & \mymethod{} (U*) & \textbf{17.2} & \textbf{33.6} & \textbf{43.3} & \textbf{57.4} & 9.6 & \textbf{24.4} & 37.1 & 58.1 & \textbf{8.5} & \textbf{26.9} & 41.3 & \textbf{60.1} & 11.7 & 26.7 & 37.5 & 50.8 \\
    \bottomrule
    \end{tabular}}
    
    \resizebox{\textwidth}{!}{
    \begin{tabular}{l|lcccccccccccccccc@{}}
    \toprule
        \multicolumn{1}{l}{} & & \multicolumn{4}{c}{Cora-ML (LT)} & \multicolumn{4}{c}{Network Science (LT)} & \multicolumn{4}{c}{Power Grid (LT)} & \multicolumn{4}{c}{Jazz (LT)} \\
        \cmidrule(lr){3-6} \cmidrule(lr){7-10} \cmidrule(lr){11-14} \cmidrule(lr){15-18}
        \multicolumn{1}{l}{Setting} & Method & 1\% & 5\% & 10\% & 20\% & 1\% & 5\% & 10\% & 20\% & 1\% & 5\% & 10\% & 20\% & 1\% & 5\% & 10\% & 20\% \\
        \midrule
        \multirow{3}{*}{\shortstack[l]{$T_0 = 100$\\$\alpha = 0.2$}}
        & \mymethod{} (C) & \textbf{20.9} & 76.7 & 84.8 & 89.4 & 4.8 & 21.9 & 28.6 & 41.9 & \textbf{7.8} & 32.4 & 54.2 & 77.4 & 1.3 & 7.4 & 70.3 & 98.0 \\
        & \mymethod{} (U) & 18.6 & 80.5 & \textbf{93.4} & 98.7 & 6.8 & 26.0 & 41.7 & 66.5 & 7.7 & 30.7 & 57.8 & 88.3 & \textbf{2.0} & \textbf{30.4} & 58.1 & \textbf{100.0} \\
        & \mymethod{} (U*) & 19.6 & \textbf{80.6} & \textbf{93.4} & 98.7 & 6.9 & 26.2 & \textbf{41.8} & 66.5 & \textbf{7.8} & 32.1 & \textbf{59.5} & \textbf{88.8} & \textbf{2.0} & \textbf{30.4} & 59.6 & \textbf{100.0} \\
        \midrule
        \multirow{3}{*}{\shortstack[l]{$T_0 = 1e^{-6}$\\$\alpha = 1$}}
        & \mymethod{} (C) & 20.7 & 76.7 & 84.8 & 89.4 & 4.9 & 22.1 & 28.6 & 41.9 & \textbf{7.8} & \textbf{32.5} & 54.2 & 77.4 & 1.3 & 7.4 & \textbf{71.2} & 98.6 \\
        & \mymethod{} (U) & 18.9 & 80.0 & 93.3 & 98.8 & 6.8 & \textbf{26.5} & 41.7 & 66.4 & 7.6 & 31.0 & 58.4 & 88.1 & \textbf{2.0} & \textbf{30.4} & 48.0 & \textbf{100.0} \\
        & \mymethod{} (U*) & 18.0 & 80.3 & 93.0 & \textbf{98.9} & \textbf{7.0} & 26.3 & 41.6 & \textbf{66.6} & \textbf{7.8} & 32.4 & \textbf{59.5} & 88.7 & \textbf{2.0} & 22.5 & 58.5 & \textbf{100.0} \\
    \bottomrule
    \end{tabular}}
    
    \resizebox{\textwidth}{!}{
    \begin{tabular}{l|lcccccccccccccccc@{}}
    \toprule
        \multicolumn{1}{l}{} & & \multicolumn{4}{c}{Cora-ML (SIS)} & \multicolumn{4}{c}{Network Science (SIS)} & \multicolumn{4}{c}{Power Grid (SIS)} & \multicolumn{4}{c}{Jazz (SIS)} \\
        \cmidrule(lr){3-6} \cmidrule(lr){7-10} \cmidrule(lr){11-14} \cmidrule(lr){15-18}
        \multicolumn{1}{l}{Setting} & Method & 1\% & 5\% & 10\% & 20\% & 1\% & 5\% & 10\% & 20\% & 1\% & 5\% & 10\% & 20\% & 1\% & 5\% & 10\% & 20\% \\
        \midrule
        \multirow{3}{*}{\shortstack[l]{$T_0 = 100$\\$\alpha = 0.2$}}
        & \mymethod{} (C) & \textbf{7.2} & 15.8 & 21.9 & 31.1 & \textbf{2.6} & 8.8 & 14.7 & 23.8 & \textbf{1.9} & \textbf{7.8} & \textbf{14.1} & 24.7 & \textbf{34.0} & 55.2 & 63.1 & 69.9 \\
        & \mymethod{} (U) & 7.1 & 15.7 & 22.4 & \textbf{32.7} & 2.4 & 8.8 & 15.4 & 26.3 & \textbf{1.9} & 7.7 & 13.9 & 25.1 & 30.4 & 56.1 & 64.5 & 71.3 \\
        & \mymethod{} (U*) & \textbf{7.2} & \textbf{15.9} & 22.4 & 32.6 & 2.4 & 8.8 & \textbf{15.5} & 26.4 & \textbf{1.9} & \textbf{7.8} & \textbf{14.1} & \textbf{25.2} & 29.9 & 56.4 & 64.3 & 71.3 \\
        \midrule
        \multirow{3}{*}{\shortstack[l]{$T_0 = 1e^{-6}$\\$\alpha = 1$}}
        & \mymethod{} (C) & \textbf{7.2} & 15.7 & 21.9 & 31.1 & 2.5 & 8.8 & 14.6 & 23.8 & \textbf{1.9} & \textbf{7.8} & \textbf{14.1} & 24.7 & 33.6 & 55.4 & 63.3 & 70.3 \\
        & \mymethod{} (U) & \textbf{7.2} & 15.7 & \textbf{22.5} & 32.5 & 2.4 & 8.8 & 15.3 & \textbf{26.5} & \textbf{1.9} & \textbf{7.8} & 14.0 & 25.1 & 30.8 & \textbf{56.7} & \textbf{64.7} & \textbf{71.6} \\
        & \mymethod{} (U*) & \textbf{7.2} & 15.8 & \textbf{22.5} & 32.4 & 2.4 & \textbf{8.9} & 15.3 & \textbf{26.5} & \textbf{1.9} & \textbf{7.8} & \textbf{14.1} & \textbf{25.2} & 32.1 & 56.6 & 64.3 & 71.2 \\
    \bottomrule
    \end{tabular}}
\end{table}

\newpage

\subsection{Influence maximization performance of \mymethod{} with standard deviations}\label{app:std}

\begin{table}[h]
    \centering
    \caption{Influence spread (\%) achieved by \mymethod{} with standard deviations.}
    \label{tab:std}
    \resizebox{\textwidth}{!}{
    \begin{tabular}{lcccccccccccccccc@{}}
    \toprule
        & \multicolumn{4}{c}{Cora-ML (IC)} & \multicolumn{4}{c}{Network Science (IC)} \\
        \cmidrule(lr){2-5} \cmidrule(lr){6-9}
        Method & 1\% & 5\% & 10\% & 20\% & 1\% & 5\% & 10\% & 20\% \\
        \midrule
        \mymethod (C) & 14.2 $\pm$ 0.2 & 33.2 $\pm$ 0.2 & 39.8 $\pm$ 0.1 & 54.7 $\pm$ 0.1 & 9.5 $\pm$ 0.1 & 24.0 $\pm$ 0.2 & 36.6 $\pm$ 0.3 & 50.3 $\pm$ 0.1 \\
        \mymethod{} (U) & 16.9 $\pm$ 0.1 & 33.3 $\pm$ 0.3 & 43.1 $\pm$ 0.3 & 57.3 $\pm$ 0.2 & 9.7 $\pm$ 0.2 & 24.0 $\pm$ 0.2 & 36.6 $\pm$ 0.6 & 57.8 $\pm$ 0.2 \\
        \mymethod{} (U*) & 17.2 $\pm$ 0.2 & 33.6 $\pm$ 0.2 & 43.3 $\pm$ 0.3 & 57.4 $\pm$ 0.0 & 9.6 $\pm$ 0.3 & 24.4 $\pm$ 0.2 & 37.1 $\pm$ 0.4 & 58.1 $\pm$ 0.2 \\
        \midrule
        & \multicolumn{4}{c}{Power Grid (IC)} & \multicolumn{4}{c}{Jazz (IC)} \\
        \cmidrule(lr){2-5} \cmidrule(lr){6-9}
        Method & 1\% & 5\% & 10\% & 20\% & 1\% & 5\% & 10\% & 20\% \\
        \midrule
        \mymethod (C) & 8.5 $\pm$ 0.1 & 26.7 $\pm$ 0.2 & 40.5 $\pm$ 0.2 & 57.6 $\pm$ 0.1 & 10.4 $\pm$ 1.6 & 27.2 $\pm$ 1.1 & 36.7 $\pm$ 0.6 & 50.0 $\pm$ 0.5 \\
        \mymethod{} (U) & 8.2 $\pm$ 0.1 & 26.2 $\pm$ 0.1 & 40.4 $\pm$ 0.3 & 59.5 $\pm$ 0.2 & 12.3 $\pm$ 0.8 & 27.6 $\pm$ 1.1 & 37.5 $\pm$ 0.6 & 51.1 $\pm$ 0.7 \\
        \mymethod{} (U*) & 8.5 $\pm$ 0.1 & 26.9 $\pm$ 0.2 & 41.3 $\pm$ 0.3 & 60.1 $\pm$ 0.1 & 11.7 $\pm$ 0.4 & 26.7 $\pm$ 0.9 & 37.5 $\pm$ 0.9 & 50.8 $\pm$ 0.6 \\
    \bottomrule
    \end{tabular}}
    
    \resizebox{\textwidth}{!}{
    \begin{tabular}{lcccccccccccccccc@{}}
    \toprule
        & \multicolumn{4}{c}{Cora-ML (LT)} & \multicolumn{4}{c}{Network Science (LT)} \\
        \cmidrule(lr){2-5} \cmidrule(lr){6-9}
        Method & 1\% & 5\% & 10\% & 20\% & 1\% & 5\% & 10\% & 20\% \\
        \midrule
        \mymethod (C) & 20.7 $\pm$ 0.4 & 76.7 $\pm$ 0.2 & 84.8 $\pm$ 0.1 & 89.4 $\pm$ 0.1 & 4.9 $\pm$ 0.3 & 22.1 $\pm$ 0.0 & 28.6 $\pm$ 0.1 & 41.9 $\pm$ 0.0 \\
        \mymethod{} (U) & 18.9 $\pm$ 0.9 & 80.0 $\pm$ 1.1 & 93.3 $\pm$ 0.4 & 98.8 $\pm$ 0.1 & 6.8 $\pm$ 0.2 & 26.5 $\pm$ 0.3 & 41.7 $\pm$ 1.0 & 66.4 $\pm$ 0.9 \\
        \mymethod{} (U*) & 18.0 $\pm$ 1.2 & 80.3 $\pm$ 0.6 & 93.0 $\pm$ 0.5 & 98.9 $\pm$ 0.1 & 7.0 $\pm$ 0.3 & 26.3 $\pm$ 0.0 & 41.6 $\pm$ 1.1 & 66.6 $\pm$ 1.0 \\
        \midrule
        & \multicolumn{4}{c}{Power Grid (LT)} & \multicolumn{4}{c}{Jazz (LT)} \\
        \cmidrule(lr){2-5} \cmidrule(lr){6-9}
        Method & 1\% & 5\% & 10\% & 20\% & 1\% & 5\% & 10\% & 20\% \\
        \midrule
        \mymethod (C) & 7.8 $\pm$ 0.1 & 32.5 $\pm$ 0.3 & 54.2 $\pm$ 0.1 & 77.4 $\pm$ 0.1 & 1.3 $\pm$ 0.0 & 7.4 $\pm$ 0.1 & 71.2 $\pm$ 1.3 & 98.6 $\pm$ 0.3 \\
        \mymethod{} (U) & 7.6 $\pm$ 0.1 & 31.0 $\pm$ 0.4 & 58.4 $\pm$ 0.5 & 88.1 $\pm$ 0.5 & 2.0 $\pm$ 0.0 & 30.4 $\pm$ 0.2 & 48.0 $\pm$ 15.3 & 100.0 $\pm$ 0.0 \\
        \mymethod{} (U*) & 7.8 $\pm$ 0.1 & 32.4 $\pm$ 0.4 & 59.5 $\pm$ 0.5 & 88.7 $\pm$ 0.6 & 2.0 $\pm$ 0.0 & 22.5 $\pm$ 10.9 & 58.5 $\pm$ 0.2 & 100.0 $\pm$ 0.0 \\
    \bottomrule
    \end{tabular}}
    
    \resizebox{\textwidth}{!}{
    \begin{tabular}{lcccccccccccccccc@{}}
    \toprule
        & \multicolumn{4}{c}{Cora-ML (SIS)} & \multicolumn{4}{c}{Network Science (SIS)} \\
        \cmidrule(lr){2-5} \cmidrule(lr){6-9}
        Method & 1\% & 5\% & 10\% & 20\% & 1\% & 5\% & 10\% & 20\% \\
        \midrule
        \mymethod (C) & 7.2 $\pm$ 0.1 & 15.7 $\pm$ 0.1 & 21.9 $\pm$ 0.1 & 31.1 $\pm$ 0.1 & 2.5 $\pm$ 0.1 & 8.8 $\pm$ 0.2 & 14.6 $\pm$ 0.0 & 23.8 $\pm$ 0.1 \\
        \mymethod{} (U) & 7.2 $\pm$ 0.1 & 15.7 $\pm$ 0.1 & 22.5 $\pm$ 0.1 & 32.5 $\pm$ 0.1 & 2.4 $\pm$ 0.1 & 8.8 $\pm$ 0.2 & 15.3 $\pm$ 0.2 & 26.5 $\pm$ 0.2 \\
        \mymethod{} (U*) & 7.2 $\pm$ 0.1 & 15.8 $\pm$ 0.1 & 22.5 $\pm$ 0.1 & 32.4 $\pm$ 0.1 & 2.4 $\pm$ 0.1 & 8.9 $\pm$ 0.2 & 15.3 $\pm$ 0.1 & 26.5 $\pm$ 0.2 \\
        \midrule
        & \multicolumn{4}{c}{Power Grid (SIS)} & \multicolumn{4}{c}{Jazz (SIS)} \\
        \cmidrule(lr){2-5} \cmidrule(lr){6-9}
        Method & 1\% & 5\% & 10\% & 20\% & 1\% & 5\% & 10\% & 20\% \\
        \midrule
        \mymethod (C) & 1.9 $\pm$ 0.0 & 7.8 $\pm$ 0.0 & 14.1 $\pm$ 0.0 & 24.7 $\pm$ 0.1 & 33.6 $\pm$ 1.4 & 55.4 $\pm$ 0.4 & 63.3 $\pm$ 0.4 & 70.3 $\pm$ 0.3 \\
        \mymethod{} (U) & 1.9 $\pm$ 0.0 & 7.8 $\pm$ 0.0 & 14.0 $\pm$ 0.1 & 25.1 $\pm$ 0.1 & 30.8 $\pm$ 2.6 & 56.7 $\pm$ 0.3 & 64.7 $\pm$ 0.3 & 71.6 $\pm$ 0.2 \\
        \mymethod{} (U*) & 1.9 $\pm$ 0.0 & 7.8 $\pm$ 0.0 & 14.1 $\pm$ 0.0 & 25.2 $\pm$ 0.1 & 32.1 $\pm$ 2.4 & 56.6 $\pm$ 0.5 & 64.3 $\pm$ 0.3 & 71.2 $\pm$ 0.5 \\
    \bottomrule
    \end{tabular}}
\end{table}

\newpage

\subsection{Comparison with DynaFLUX}\label{app:dynaflux}

\begin{table}[h]
    \centering
    \caption{Influence spread (\%) achieved by \mymethod{} and DynaFLUX.}
    \label{tab:dynaflux}
    \resizebox{\textwidth}{!}{
    \begingroup \setlength{\tabcolsep}{3pt}
    \begin{tabular}{@{}lcccccccccccccccccccccccc@{}}
    \toprule
        & \multicolumn{4}{c}{Cora-ML (IC)} & \multicolumn{4}{c}{Network Science (IC)} & \multicolumn{4}{c}{Power Grid (IC)} & \multicolumn{4}{c}{Jazz (IC)} & \multicolumn{4}{c}{Synthetic (IC)} & \multicolumn{4}{c}{Digg (IC)} \\
        \cmidrule(lr){2-5} \cmidrule(lr){6-9} \cmidrule(lr){10-13} \cmidrule(lr){14-17} \cmidrule(lr){18-21} \cmidrule(lr){22-25}
        Method & 1\% & 5\% & 10\% & 20\% & 1\% & 5\% & 10\% & 20\% & 1\% & 5\% & 10\% & 20\% & 1\% & 5\% & 10\% & 20\% & 1\% & 5\% & 10\% & 20\% & 1\% & 5\% & 10\% & 20\% \\
        \midrule
        DynaFLUX   & 35.3 & 46.8 & 55.5 & 65.4 &  7.8 & 21.1 & 29.6 & 41.7 &  9.0 & 27.4 & 41.3 & 58.6 & 26.0 & 37.3 & 41.7 & 56.4 & \textbf{14.4} & \textbf{30.4} & 41.1 & 54.4 & 60.3 & 66.2 & 74.8 & \textbf{84.6} \\
        DynaFLUX+  & \textbf{38.4} & \textbf{47.4} & \textbf{56.7} & \textbf{68.2} & 9.3 & 23.4 & 32.9 & 45.1 & \textbf{9.1} & \textbf{29.0} & \textbf{43.1} & \textbf{61.6} & \textbf{26.3} & \textbf{39.7} & \textbf{45.2} & \textbf{60.8} & 14.1 & 30.3 & \textbf{41.4} & \textbf{54.5} & \textbf{60.8} & \textbf{68.6} & \textbf{74.1} & 78.4 \\
        \midrule
        \mymethod{} (C) & 14.2 & 33.2 & 39.8 & 54.7 & 9.5 & 24.0 & 36.6 & 50.3 & 8.5 & 26.7 & 40.5 & 57.6 & 10.4 & 27.2 & 36.7 & 50.0 & 11.9 & 28.3 & 38.5 & 52.2 & 60.3 & 66.3 & 72.3 & 80.6 \\
        \mymethod{} (U) & 16.9 & 33.3 & 43.1 & 57.3 & \textbf{9.7} & 24.0 & 36.6 & 57.8 & 8.2 & 26.2 & 40.4 & 59.5 & 12.3 & 27.6 & 37.5 & 51.1 & 11.6 & 28.4 & 39.4 & 53.8 & 60.4 & 66.5 & 72.3 & 80.6 \\
        \mymethod{} (U*) & 17.2 & 33.6 & 43.3 & 57.4 & 9.6 & \textbf{24.4} & \textbf{37.1} & \textbf{58.1} & 8.5 & 26.9 & 41.3 & 60.1 & 11.7 & 26.7 & 37.5 & 50.8 & 12.0 & 28.3 & 39.4 & 53.8 & 60.3 & 66.4 & 72.5 & 81.2 \\
    \bottomrule
    \end{tabular}\endgroup}
    
    \resizebox{\textwidth}{!}{
    \begingroup \setlength{\tabcolsep}{3pt}
    \begin{tabular}{@{}lcccccccccccccccccccccccc@{}}
        \toprule
        & \multicolumn{4}{c}{Cora-ML (LT)} & \multicolumn{4}{c}{Network Science (LT)} & \multicolumn{4}{c}{Power Grid (LT)} & \multicolumn{4}{c}{Jazz (LT)} & \multicolumn{4}{c}{Synthetic (LT)} & \multicolumn{4}{c}{Digg (LT)} \\
        \cmidrule(lr){2-5} \cmidrule(lr){6-9} \cmidrule(lr){10-13} \cmidrule(lr){14-17} \cmidrule(lr){18-21} \cmidrule(lr){22-25}
        Method & 1\% & 5\% & 10\% & 20\% & 1\% & 5\% & 10\% & 20\% & 1\% & 5\% & 10\% & 20\% & 1\% & 5\% & 10\% & 20\% & 1\% & 5\% & 10\% & 20\% & 1\% & 5\% & 10\% & 20\% \\
        \midrule
        DynaFLUX   & 15.1 & 73.8 & 89.5 & 94.4 & 5.0 & 20.3 & 30.8 & 45.0 & 8.3 & 31.5 & 58.3 & 83.2 & 1.5 & 7.1 & 36.9 & 99.9 & 1.0 & 5.2 & 11.2 & 99.9 & 79.6 & 85.4 & \textbf{91.9} & \textbf{97.1} \\
        DynaFLUX+  & 15.3 & 73.7 & 90.0 & 96.2 & 6.0 & 23.7 & 34.8 & 49.5 & \textbf{8.7} & \textbf{34.9} & \textbf{62.7} & 88.4 & 1.6 & 7.6 & 14.2 & 99.9 & 1.1 & 5.3 & 11.4 & 99.9 & \textbf{80.9} & \textbf{87.6} & 90.3 & 91.7 \\
        \midrule
        \mymethod{} (C) & \textbf{20.7} & 76.7 & 84.8 & 89.4 & 4.9 & 22.1 & 28.6 & 41.9 & 7.8 & 32.5 & 54.2 & 77.4 & 1.3 & 7.4 & \textbf{71.2} & 98.6 & 1.1 & 6.5 & 15.9 & \textbf{100.0} & 12.3 & 46.4 & 85.0 & 90.6 \\
        \mymethod{} (U) & 18.9 & 80.0 & \textbf{93.3} & 98.8 & 6.8 & \textbf{26.5} & \textbf{41.7} & 66.4 & 7.6 & 31.0 & 58.4 & 88.1 & \textbf{2.0} & \textbf{30.4} & 48.0 & \textbf{100.0} & \textbf{1.4} & 7.0 & 15.9 & \textbf{100.0} & 15.4 & 81.9 & 85.2 & 90.9 \\
        \mymethod{} (U*) & 18.0 & \textbf{80.3} & 93.0 & \textbf{98.9} & \textbf{7.0} & 26.3 & 41.6 & \textbf{66.6} & 7.8 & 32.4 & 59.5 & \textbf{88.7} & \textbf{2.0} & 22.5 & 58.5 & \textbf{100.0} & \textbf{1.4} & \textbf{7.1} & \textbf{16.4} & \textbf{100.0} & 18.6 & 82.4 & 86.3 & 91.8 \\
        \bottomrule
    \end{tabular}\endgroup}

    \resizebox{\textwidth}{!}{
    \begingroup \setlength{\tabcolsep}{3pt}
    \begin{tabular}{@{}lcccccccccccccccccccccccc@{}}
        \toprule
        & \multicolumn{4}{c}{Cora-ML (SIS)} & \multicolumn{4}{c}{Network Science (SIS)} & \multicolumn{4}{c}{Power Grid (SIS)} & \multicolumn{4}{c}{Jazz (SIS)} & \multicolumn{4}{c}{Synthetic (SIS)} & \multicolumn{4}{c}{Digg (SIS)} \\
        \cmidrule(lr){2-5} \cmidrule(lr){6-9} \cmidrule(lr){10-13} \cmidrule(lr){14-17} \cmidrule(lr){18-21} \cmidrule(lr){22-25}
        Method & 1\% & 5\% & 10\% & 20\% & 1\% & 5\% & 10\% & 20\% & 1\% & 5\% & 10\% & 20\% & 1\% & 5\% & 10\% & 20\% & 1\% & 5\% & 10\% & 20\% & 1\% & 5\% & 10\% & 20\% \\
        \midrule
        DynaFLUX  & 6.9 & 15.2 & 20.8 & 29.9 & 2.3 & 7.0 & 12.4 & 22.7 & 1.8 & 6.8 & 12.2 & 22.3 & 30.6 & 56.4 & 64.5 & 70.3 & 3.0 & 12.6 & 22.2 & 37.0 & 16.5 & 18.0 & 21.1 & 28.5 \\
        DynaFLUX+ & \textbf{7.3} & \textbf{15.9} & \textbf{22.5} & 31.9 & \textbf{2.8} & 8.8 & 14.1 & 24.2 & \textbf{2.3} & 7.6 & 13.7 & \textbf{25.4} & \textbf{35.8} & \textbf{58.9} & \textbf{64.7} & 68.6 & 3.2 & 13.5 & 23.6 & 38.4 & \textbf{17.1} & \textbf{20.4} & 23.7 & 31.7 \\
        \midrule
        \mymethod{} (C) & 7.2 & 15.7 & 21.9 & 31.1 & 2.5 & 8.8 & 14.6 & 23.8 & 1.9 & \textbf{7.8} & \textbf{14.1} & 24.7 & 33.6 & 55.4 & 63.3 & 70.3 & \textbf{3.5} & \textbf{14.6} & 24.8 & 39.7 & 16.5 & 20.3 & \textbf{24.9} & 33.5 \\
        \mymethod{} (U) & 7.2 & 15.7 & \textbf{22.5} & \textbf{32.5} & 2.4 & 8.8 & \textbf{15.3} & \textbf{26.5} & 1.9 & \textbf{7.8} & 14.0 & 25.1 & 30.8 & 56.7 & \textbf{64.7} & \textbf{71.6} & 3.4 & 14.3 & 24.7 & 40.2 & 16.5 & 20.3 & \textbf{24.9} & 33.5 \\
        \mymethod{} (U*) & 7.2 & 15.8 & \textbf{22.5} & 32.4 & 2.4 & \textbf{8.9} & \textbf{15.3} & \textbf{26.5} & 1.9 & \textbf{7.8} & \textbf{14.1} & 25.2 & 32.1 & 56.6 & 64.3 & 71.2 & \textbf{3.5} & \textbf{14.6} & \textbf{24.9} & \textbf{40.4} & 16.5 & \textbf{20.4} & \textbf{24.9} & \textbf{33.6} \\
        \bottomrule
    \end{tabular}\endgroup}
\end{table}

For a more comprehensive comparison and to show the competitiveness of \mymethod{}, we include DynaFLUX~\citep{zhang2026dynaflux} in Table~\ref{tab:dynaflux}, which incorporates more information into its model by observing the node states at each diffusion step and requires access to the underlying diffusion models to run Monte-Carlo simulations to compute the reward during reinforcement learning training.

\subsection{Results on Digg using different search parameters}\label{app:digg}

\begin{table}[h]
    \centering
    \caption{Influence spread (\%) achieved by \mymethod{} on Digg using different search parameters.}
    \label{tab:digg}
    \resizebox{\textwidth}{!}{
    \begin{tabular}{l|lcccccccccccc}
    \toprule
        \multicolumn{1}{l}{} & & \multicolumn{4}{c}{Digg (IC)} & \multicolumn{4}{c}{Digg (LT)} & \multicolumn{4}{c}{Digg (SIS)} \\
        \cmidrule(lr){3-6} \cmidrule(lr){7-10} \cmidrule(lr){11-14}
        \multicolumn{1}{l}{Setting} & Method & 1\% & 5\% & 10\% & 20\% & 1\% & 5\% & 10\% & 20\% & 1\% & 5\% & 10\% & 20\% \\
        \midrule
        \multirow{3}{*}{\shortstack[l]{\# of swaps = 1\\\# of steps = 50,000}}
        & \mymethod{} (C) & 60.6 & 68.3 & 74.8 & 82.3 & 11.3 & 22.5 & 35.2 & 91.8 & 16.5 & 20.5 & 25.0 & 33.5 \\
        & \mymethod{} (U) & 60.7 & 68.4 & 74.8 & 82.3 & 11.1 & 21.7 & 33.8 & 91.2 & 16.5 & 20.5 & 24.9 & 33.6 \\
        & \mymethod{} (U*) & 60.4 & 68.7 & 76.1 & 84.0 & 18.8 & 82.7 & 87.4 & 93.8 & 16.5 & 20.7 & 25.1 & 33.8 \\
        \midrule
        \multirow{3}{*}{\shortstack[l]{\# of swaps = 20\\\# of steps = 20,000}}
        & \mymethod{} (C) & 60.3 & 66.3 & 72.3 & 80.6 & 12.3 & 46.4 & 85.0 & 90.6 & 16.5 & 20.3 & 24.9 & 33.5 \\
        & \mymethod{} (U) & 60.4 & 66.5 & 72.3 & 80.6 & 15.4 & 81.9 & 85.2 & 90.9 & 16.5 & 20.3 & 24.9 & 33.5 \\
        & \mymethod{} (U*) & 60.3 & 66.4 & 72.5 & 81.2 & 18.6 & 82.4 & 86.3 & 91.9 & 16.5 & 20.4 & 24.9 & 33.6 \\
    \bottomrule
    \end{tabular}}
\end{table}

\newpage

\subsection{Full results on data efficiency}\label{app:data_efficiency_full}

\begin{table}[h]
    \centering
    \caption{Influence spread (\%) achieved by all \mymethod{} variants with varying fractions of training data.}
    \label{tab:data_efficiency_full}
    \resizebox{\textwidth}{!}{
    \begin{tabular}{c|lcccccccccccccccc@{}}
    \toprule
        \multicolumn{1}{l}{} & & \multicolumn{4}{c}{Cora-ML (IC)} & \multicolumn{4}{c}{Network Science (IC)} & \multicolumn{4}{c}{Power Grid (IC)} & \multicolumn{4}{c}{Jazz (IC)} \\
        \cmidrule(lr){3-6} \cmidrule(lr){7-10} \cmidrule(lr){11-14} \cmidrule(lr){15-18}
        \multicolumn{1}{c}{Training data} & Method & 1\% & 5\% & 10\% & 20\% & 1\% & 5\% & 10\% & 20\% & 1\% & 5\% & 10\% & 20\% & 1\% & 5\% & 10\% & 20\% \\
        \midrule
        \multirow{3}{*}{1\%}
        & \mymethod{} (C) & 14.2 & 32.2 & 39.5 & 54.4 & 9.4 & 24.1 & 36.9 & 50.1 & 8.2 & 26.3 & 40.2 & 57.5 & 10.1 & 27.1 & 36.4 & 49.4 \\
        & \mymethod{} (U) & 16.8 & 32.2 & 42.8 & 57.2 & 9.7 & 23.8 & 37.0 & 58.0 & 7.9 & 25.8 & 40.2 & 59.3 & 12.1 & 27.5 & 37.5 & 49.3 \\
        & \mymethod{} (U*) & 16.8 & 32.4 & 43.1 & 57.2 & 9.7 & 24.4 & 37.3 & 58.0 & 8.2 & 26.4 & 41.1 & 59.8 & 12.1 & 27.2 & 37.4 & 49.7 \\
        \midrule
        \multirow{3}{*}{5\%}
        & \mymethod{} (C) & 14.1 & 32.5 & 39.8 & 54.4 & 9.4 & 24.1 & 36.8 & 50.2 & 8.2 & 26.4 & 40.3 & 57.4 & 10.4 & 27.2 & 36.4 & 49.7 \\
        & \mymethod{} (U) &  16.9 & 32.5 & 43.2 & 56.8 & 9.7 & 24.0 & 36.8 & 58.0 & 8.2 & 25.8 & 40.4 & 59.5 & 12.3 & 27.5 & 37.1 & 50.2 \\
        & \mymethod{} (U*) & 16.7 & 32.7 & 43.2 & 56.8 & 9.7 & 24.4 & 37.3 & 58.2 & 8.2 & 26.7 & 41.1 & 59.8 & 11.6 & 27.2 & 36.7 & 50.5 \\
        \midrule
        \multirow{3}{*}{20\%}
        & \mymethod{} (C) & 14.2 & 32.6 & 39.8 & 54.5 & 9.6 & 24.1 & 36.9 & 50.3 & 8.2 & 26.6 & 40.5 & 57.6 & 9.3 & 27.4 & 36.3 & 50.3 \\
        & \mymethod{} (U) & 16.9 & 32.4 & 43.1 & 57.0 & 9.7 & 23.9 & 37.1 & 58.0 & 7.9 & 25.9 & 40.6 & 59.4 & 11.7 & 26.6 & 36.3 & 50.3 \\
        & \mymethod{} (U*) & 17.0 & 32.9 & 43.1 & 57.1 & 9.5 & 24.3 & 37.3 & 58.2 & 8.3 & 26.7 & 41.4 & 59.9 & 12.0 & 27.6 & 36.9 & 50.5 \\
        \midrule
        \multirow{3}{*}{100\%}
        & \mymethod{} (C) & 14.2 & 33.2 & 39.8 & 54.7 & 9.5 & 24.0 & 36.6 & 50.3 & 8.5 & 26.7 & 40.5 & 57.6 & 10.4 & 27.2 & 36.7 & 50.0 \\
        & \mymethod{} (U) & 16.9 & 33.3 & 43.1 & 57.3 & 9.7 & 24.0 & 36.6 & 57.8 & 8.2 & 26.2 & 40.4 & 59.5 & 12.3 & 27.6 & 37.5 & 51.1 \\
        & \mymethod{} (U*) & 17.2 & 33.6 & 43.3 & 57.4 & 9.6 & 24.4 & 37.1 & 58.1 & 8.5 & 26.9 & 41.3 & 60.1 & 11.7 & 26.7 & 37.5 & 50.8 \\
    \bottomrule
    \end{tabular}}
    
    \resizebox{\textwidth}{!}{
    \begin{tabular}{c|lcccccccccccccccc@{}}
    \toprule
        \multicolumn{1}{l}{} & & \multicolumn{4}{c}{Cora-ML (LT)} & \multicolumn{4}{c}{Network Science (LT)} & \multicolumn{4}{c}{Power Grid (LT)} & \multicolumn{4}{c}{Jazz (LT)} \\
        \cmidrule(lr){3-6} \cmidrule(lr){7-10} \cmidrule(lr){11-14} \cmidrule(lr){15-18}
        \multicolumn{1}{c}{Training data} & Method & 1\% & 5\% & 10\% & 20\% & 1\% & 5\% & 10\% & 20\% & 1\% & 5\% & 10\% & 20\% & 1\% & 5\% & 10\% & 20\% \\
        \midrule
        \multirow{3}{*}{1\%}
        & \mymethod{} (C) & 20.9 & 76.9 & 84.8 & 89.3 & 4.9 & 21.9 & 28.5 & 41.8 & 7.8 & 32.5 & 54.2 & 77.3 & 1.3 & 7.5 & 72.0 & 98.1 \\
        & \mymethod{} (U) & 18.1 & 80.8 & 93.4 & 98.8 & 6.9 & 26.1 & 42.4 & 66.8 & 7.6 & 30.9 & 58.0 & 88.4 & 2.0 & 23.0 & 58.7 & 100.0 \\
        & \mymethod{} (U*) & 18.6 & 81.2 & 93.1 & 98.8 & 6.9 & 26.1 & 42.2 & 66.8 & 7.8 & 32.5 & 59.3 & 88.7 & 2.0 & 22.9 & 45.6 & 100.0 \\
        \midrule
        \multirow{3}{*}{5\%}
        & \mymethod{} (C) & 19.9 & 76.8 & 84.8 & 89.4 & 4.9 & 21.8 & 28.5 & 41.9 & 7.8 & 32.2 & 54.2 & 77.4 & 1.3 & 7.4 & 70.4 & 97.9 \\
        & \mymethod{} (U) & 19.0 & 80.7 & 92.8 & 98.9 & 6.9 & 25.9 & 41.8 & 66.2 & 7.6 & 30.6 & 58.3 & 88.3 & 2.0 & 22.7 & 56.8 & 100.0 \\
        & \mymethod{} (U*) & 18.4 & 80.5 & 92.9 & 98.7 & 7.1 & 26.2 & 41.8 & 66.4 & 7.8 & 32.1 & 59.4 & 88.7 & 2.0 & 22.5 & 52.1 & 100.0 \\
        \midrule
        \multirow{3}{*}{20\%}
        & \mymethod{} (C) & 20.7 & 76.9 & 84.8 & 89.4 & 4.7 & 21.9 & 28.6 & 41.9 & 7.8 & 32.3 & 54.1 & 77.4 & 1.4 & 7.4 & 70.5 & 98.3 \\
        & \mymethod{} (U) & 19.3 & 80.7 & 93.0 & 98.7 & 6.8 & 26.3 & 42.3 & 67.0 & 7.6 & 30.7 & 58.2 & 88.1 & 2.0 & 30.4 & 54.5 & 100.0 \\
        & \mymethod{} (U*) & 18.5 & 80.7 & 93.2 & 98.7 & 6.7 & 26.4 & 42.5 & 66.7 & 7.8 & 32.1 & 59.5 & 88.8 & 2.0 & 27.0 & 56.7 & 100.0 \\
        \midrule
        \multirow{3}{*}{100\%}
        & \mymethod{} (C) & 20.7 & 76.7 & 84.8 & 89.4 & 4.9 & 22.1 & 28.6 & 41.9 & 7.8 & 32.5 & 54.2 & 77.4 & 1.3 & 7.4 & 71.2 & 98.6 \\
        & \mymethod{} (U) & 18.9 & 80.0 & 93.3 & 98.8 & 6.8 & 26.5 & 41.7 & 66.4 & 7.6 & 31.0 & 58.4 & 88.1 & 2.0 & 30.4 & 48.0 & 100.0 \\
        & \mymethod{} (U*) & 18.0 & 80.3 & 93.0 & 98.9 & 7.0 & 26.3 & 41.6 & 66.6 & 7.8 & 32.4 & 59.5 & 88.7 & 2.0 & 22.5 & 58.5 & 100.0 \\
    \bottomrule
    \end{tabular}}

    \resizebox{\textwidth}{!}{
    \begin{tabular}{c|lcccccccccccccccc@{}}
    \toprule
        \multicolumn{1}{l}{} & & \multicolumn{4}{c}{Cora-ML (SIS)} & \multicolumn{4}{c}{Network Science (SIS)} & \multicolumn{4}{c}{Power Grid (SIS)} & \multicolumn{4}{c}{Jazz (SIS)} \\
        \cmidrule(lr){3-6} \cmidrule(lr){7-10} \cmidrule(lr){11-14} \cmidrule(lr){15-18}
        \multicolumn{1}{c}{Training data} & Method & 1\% & 5\% & 10\% & 20\% & 1\% & 5\% & 10\% & 20\% & 1\% & 5\% & 10\% & 20\% & 1\% & 5\% & 10\% & 20\% \\
        \midrule
        \multirow{3}{*}{1\%}
        & \mymethod{} (C) & 7.1 & 15.6 & 21.5 & 30.5 & 2.5 & 8.7 & 14.5 & 23.5 & 1.8 & 7.5 & 13.4 & 24.1 & 35.1 & 55.1 & 62.9 & 70.3 \\
        & \mymethod{} (U) & 7.0 & 15.5 & 21.9 & 31.6 & 2.4 & 9.0 & 14.6 & 25.4 & 1.8 & 7.4 & 13.4 & 24.3 & 35.2 & 56.1 & 64.6 & 71.7 \\
        & \mymethod{} (U*) & 7.1 & 15.7 & 22.0 & 31.6 & 2.5 & 9.0 & 14.7 & 25.4 & 1.8 & 7.4 & 13.4 & 24.5 & 35.0 & 56.3 & 64.5 & 71.5 \\
        \midrule
        \multirow{3}{*}{5\%}
        & \mymethod{} (C) & 7.2 & 15.5 & 21.8 & 31.0 & 2.5 & 8.8 & 14.3 & 23.7 & 1.8 & 7.7 & 13.6 & 24.4 & 33.1 & 55.0 & 62.9 & 70.2 \\
        & \mymethod{} (U) & 7.0 & 15.4 & 22.5 & 32.3 & 2.3 & 9.1 & 15.1 & 25.9 & 1.8 & 7.5 & 13.5 & 24.7 & 32.8 & 56.6 & 64.7 & 71.4 \\
        & \mymethod{} (U*) & 7.2 & 15.5 & 22.6 & 32.3 & 2.3 & 9.1 & 15.2 & 25.9 & 1.8 & 7.6 & 13.6 & 24.9 & 31.8 & 56.8 & 64.5 & 71.4 \\
        \midrule
        \multirow{3}{*}{20\%}
        & \mymethod{} (C) & 7.2 & 15.8 & 21.8 & 31.0 & 2.5 & 8.9 & 14.6 & 23.8 & 1.8 & 7.7 & 14.0 & 24.6 & 33.8 & 54.8 & 63.4 & 70.2 \\
        & \mymethod{} (U) & 7.1 & 15.8 & 22.4 & 32.4 & 2.4 & 8.8 & 15.2 & 26.0 & 1.8 & 7.6 & 13.8 & 25.0 & 28.9 & 56.6 & 64.9 & 71.2 \\
        & \mymethod{} (U*) & 7.2 & 15.9 & 22.4 & 32.5 & 2.4 & 8.8 & 15.1 & 26.0 & 1.9 & 7.8 & 14.0 & 25.1 & 29.9 & 56.9 & 64.6 & 71.5 \\
        \midrule
        \multirow{3}{*}{100\%}
        & \mymethod{} (C) & 7.2 & 15.7 & 21.9 & 31.1 & 2.5 & 8.8 & 14.6 & 23.8 & 1.9 & 7.8 & 14.1 & 24.7 & 33.6 & 55.4 & 63.3 & 70.3 \\
        & \mymethod{} (U) & 7.2 & 15.7 & 22.5 & 32.5 & 2.4 & 8.8 & 15.3 & 26.5 & 1.9 & 7.8 & 14.0 & 25.1 & 30.8 & 56.7 & 64.7 & 71.6 \\
        & \mymethod{} (U*) & 7.2 & 15.8 & 22.5 & 32.4 & 2.4 & 8.9 & 15.3 & 26.5 & 1.9 & 7.8 & 14.1 & 25.2 & 32.1 & 56.6 & 64.3 & 71.2 \\
    \bottomrule
    \end{tabular}}
\end{table}

\newpage

\subsection{Full results on the effect of node embedding initialization strategy on seed set quality}\label{app:init_full}

\begin{table}[h]
    \centering
    \caption{Influence spread (\%) achieved by all \mymethod{} variants with different node embedding initialization strategies and post-training embedding update rules.}
    \label{tab:init_full}
    \resizebox{\textwidth}{!}{
    \begin{tabular}{l|lcccccccccccccccc@{}}
    \toprule
        \multicolumn{1}{l}{} & & \multicolumn{4}{c}{Cora-ML (IC)} & \multicolumn{4}{c}{Network Science (IC)} & \multicolumn{4}{c}{Power Grid (IC)} & \multicolumn{4}{c}{Jazz (IC)} \\
        \cmidrule(lr){3-6} \cmidrule(lr){7-10} \cmidrule(lr){11-14} \cmidrule(lr){15-18}
        \multicolumn{1}{l}{Setting} & Method & 1\% & 5\% & 10\% & 20\% & 1\% & 5\% & 10\% & 20\% & 1\% & 5\% & 10\% & 20\% & 1\% & 5\% & 10\% & 20\% \\
        \midrule
        \multirow{3}{*}{\shortstack[l]{Random\\No update}}
        & \mymethod{} (C) & 14.0 & 32.3 & 39.1 & 53.9 & 9.2 & 23.9 & 36.7 & 49.6 & 8.1 & 26.2 & 40.0 & 56.8 & 9.8 & 26.9 & 35.5 & 49.3 \\
        & \mymethod{} (U) & 16.4 & 32.4 & 42.4 & 56.3 & 9.4 & 23.7 & 36.5 & 56.9 & 7.9 & 25.7 & 39.8 & 58.7 & 10.4 & 25.3 & 36.1 & 50.5 \\
        & \mymethod{} (U*) & 16.5 & 32.8 & 42.3 & 56.1 & 9.5 & 23.9 & 36.6 & 57.1 & 8.0 & 26.2 & 40.6 & 59.0 & 9.9 & 25.8 & 36.7 & 50.6 \\
        \midrule
        \multirow{3}{*}{\shortstack[l]{Random\\Weighted avg}}
        & \mymethod{} (C) & 14.0 & 32.2 & 39.4 & 53.8 & 9.2 & 23.9 & 36.5 & 49.7 & 8.2 & 26.2 & 39.8 & 56.6 & 10.3 & 26.5 & 35.9 & 49.3 \\
        & \mymethod{} (U) & 14.8 & 32.4 & 38.9 & 53.0 & 9.2 & 23.7 & 36.0 & 49.6 & 7.9 & 25.4 & 38.5 & 55.0 & 10.6 & 26.8 & 36.3 & 49.4 \\
        & \mymethod{} (U*) & 14.8 & 32.5 & 39.3 & 53.8 & 9.2 & 24.0 & 36.6 & 49.7 & 8.3 & 26.4 & 40.0 & 56.8 & 11.1 & 26.9 & 35.6 & 49.1 \\
        \midrule
        \multirow{3}{*}{\shortstack[l]{Random\\Most similar}}
        & \mymethod{} (C) & 13.9 & 32.5 & 39.5 & 54.0 & 9.3 & 24.0 & 36.6 & 49.7 & 8.2 & 26.2 & 39.8 & 56.7 & 8.7 & 26.6 & 35.5 & 50.0 \\
        & \mymethod{} (U) & 16.7 & 32.5 & 42.7 & 56.2 & 9.3 & 24.0 & 36.6 & 56.7 & 7.8 & 25.5 & 39.8 & 58.8 & 9.7 & 26.8 & 36.6 & 50.5 \\
        & \mymethod{} (U*) & 16.4 & 32.8 & 42.8 & 56.3 & 9.4 & 24.4 & 36.8 & 56.4 & 8.2 & 26.3 & 40.5 & 59.1 & 9.7 & 26.9 & 36.4 & 50.1 \\
        \midrule
        \multirow{3}{*}{\shortstack[l]{Random\\Global mean}}
        & \mymethod{} (C) & 14.2 & 32.2 & 39.3 & 53.8 & 9.3 & 23.9 & 36.5 & 49.8 & 8.1 & 26.2 & 39.8 & 56.4 & 10.1 & 26.3 & 36.4 & 49.1 \\
        & \mymethod{} (U) & 14.4 & 32.3 & 38.6 & 53.3 & 9.3 & 23.7 & 36.1 & 49.4 & 8.0 & 25.3 & 38.8 & 54.7 & 10.4 & 26.5 & 35.9 & 49.4 \\
        & \mymethod{} (U*) & 14.6 & 32.6 & 39.4 & 53.9 & 9.4 & 24.0 & 36.5 & 49.8 & 8.2 & 26.3 & 40.0 & 56.5 & 9.9 & 26.8 & 36.1 & 49.3 \\
        \midrule
        \multirow{3}{*}{\shortstack[l]{Anchored\\No update}}
        & \mymethod{} (C) & 14.2 & 33.2 & 39.8 & 54.7 & 9.5 & 24.0 & 36.6 & 50.3 & 8.5 & 26.7 & 40.5 & 57.6 & 10.4 & 27.2 & 36.7 & 50.0 \\
        & \mymethod{} (U) & 16.9 & 33.3 & 43.1 & 57.3 & 9.7 & 24.0 & 36.6 & 57.8 & 8.2 & 26.2 & 40.4 & 59.5 & 12.3 & 27.6 & 37.5 & 51.1 \\
        & \mymethod{} (U*) & 17.2 & 33.6 & 43.3 & 57.4 & 9.6 & 24.4 & 37.1 & 58.1 & 8.5 & 26.9 & 41.3 & 60.1 & 11.7 & 26.7 & 37.5 & 50.8 \\
    \bottomrule
    \end{tabular}}
    
    \resizebox{\textwidth}{!}{
    \begin{tabular}{l|lcccccccccccccccc@{}}
    \toprule
        \multicolumn{1}{l}{} & & \multicolumn{4}{c}{Cora-ML (LT)} & \multicolumn{4}{c}{Network Science (LT)} & \multicolumn{4}{c}{Power Grid (LT)} & \multicolumn{4}{c}{Jazz (LT)} \\
        \cmidrule(lr){3-6} \cmidrule(lr){7-10} \cmidrule(lr){11-14} \cmidrule(lr){15-18}
        \multicolumn{1}{l}{Setting} & Method & 1\% & 5\% & 10\% & 20\% & 1\% & 5\% & 10\% & 20\% & 1\% & 5\% & 10\% & 20\% & 1\% & 5\% & 10\% & 20\% \\
        \midrule
        \multirow{3}{*}{\shortstack[l]{Random\\No update}}
        & \mymethod{} (C) & 16.8 & 75.5 & 84.3 & 89.3 & 4.7 & 21.0 & 28.6 & 41.9 & 7.3 & 30.8 & 53.4 & 76.8 & 1.4 & 6.9 & 69.8 & 88.9 \\
        & \mymethod{} (U) & 15.3 & 79.1 & 90.9 & 97.6 & 5.6 & 22.0 & 35.2 & 59.4 & 6.6 & 26.7 & 52.7 & 83.1 & 1.8 & 7.7 & 24.0 & 99.5 \\
        & \mymethod{} (U*) & 15.0 & 79.1 & 91.1 & 97.6 & 5.8 & 21.9 & 35.0 & 59.3 & 6.8 & 28.2 & 54.2 & 83.6 & 1.8 & 7.7 & 24.1 & 99.0 \\
        \midrule
        \multirow{3}{*}{\shortstack[l]{Random\\Weighted avg}}
        & \mymethod{} (C) & 17.3 & 75.5 & 84.3 & 89.3 & 4.7 & 20.9 & 28.6 & 41.9 & 7.3 & 30.9 & 53.5 & 77.1 & 1.3 & 7.1 & 67.9 & 85.6 \\
        & \mymethod{} (U) & 16.9 & 74.1 & 84.0 & 89.3 & 4.7 & 21.0 & 29.8 & 43.0 & 7.0 & 29.2 & 51.3 & 75.8 & 1.9 & 7.1 & 68.6 & 94.9 \\
        & \mymethod{} (U*) & 17.6 & 75.6 & 84.4 & 89.2 & 4.7 & 20.9 & 28.8 & 42.0 & 7.3 & 30.9 & 53.4 & 77.1 & 1.9 & 7.1 & 67.7 & 94.8 \\
        \midrule
        \multirow{3}{*}{\shortstack[l]{Random\\Most similar}}
        & \mymethod{} (C) & 16.3 & 74.9 & 84.3 & 89.3 & 4.8 & 20.9 & 28.5 & 41.9 & 7.3 & 30.7 & 53.5 & 77.1 & 1.4 & 7.0 & 67.9 & 88.9 \\
        & \mymethod{} (U) & 15.3 & 78.0 & 90.9 & 97.4 & 6.6 & 22.0 & 37.6 & 61.7 & 7.0 & 27.3 & 54.0 & 85.0 & 2.0 & 13.1 & 24.5 & 98.8 \\
        & \mymethod{} (U*) & 15.6 & 78.8 & 90.5 & 97.6 & 6.5 & 22.3 & 37.6 & 61.4 & 7.1 & 28.7 & 54.8 & 85.9 & 2.0 & 13.1 & 25.0 & 99.5 \\
        \midrule
        \multirow{3}{*}{\shortstack[l]{Random\\Global mean}}
        & \mymethod{} (C) & 16.9 & 74.7 & 84.3 & 89.3 & 4.8 & 21.1 & 28.5 & 41.8 & 7.3 & 31.1 & 53.3 & 76.8 & 1.3 & 7.1 & 68.8 & 86.2 \\
        & \mymethod{} (U) & 16.1 & 73.7 & 84.4 & 89.6 & 4.9 & 21.1 & 28.9 & 42.3 & 7.1 & 29.8 & 51.1 & 75.6 & 1.9 & 7.1 & 66.4 & 99.5 \\
        & \mymethod{} (U*) & 16.9 & 74.9 & 84.4 & 89.3 & 4.9 & 21.1 & 28.5 & 41.9 & 7.3 & 31.0 & 53.2 & 76.8 & 1.9 & 7.1 & 64.9 & 99.5 \\
        \midrule
        \multirow{3}{*}{\shortstack[l]{Anchored\\No update}}
        & \mymethod{} (C) & 20.7 & 76.7 & 84.8 & 89.4 & 4.9 & 22.1 & 28.6 & 41.9 & 7.8 & 32.5 & 54.2 & 77.4 & 1.3 & 7.4 & 71.2 & 98.6 \\
        & \mymethod{} (U) & 18.9 & 80.0 & 93.3 & 98.8 & 6.8 & 26.5 & 41.7 & 66.4 & 7.6 & 31.0 & 58.4 & 88.1 & 2.0 & 30.4 & 48.0 & 100.0 \\
        & \mymethod{} (U*) & 18.0 & 80.3 & 93.0 & 98.9 & 7.0 & 26.3 & 41.6 & 66.6 & 7.8 & 32.4 & 59.5 & 88.7 & 2.0 & 22.5 & 58.5 & 100.0 \\
    \bottomrule
    \end{tabular}}
    
    \resizebox{\textwidth}{!}{
    \begin{tabular}{l|lcccccccccccccccc@{}}
    \toprule
        \multicolumn{1}{l}{} & & \multicolumn{4}{c}{Cora-ML (SIS)} & \multicolumn{4}{c}{Network Science (SIS)} & \multicolumn{4}{c}{Power Grid (SIS)} & \multicolumn{4}{c}{Jazz (SIS)} \\
        \cmidrule(lr){3-6} \cmidrule(lr){7-10} \cmidrule(lr){11-14} \cmidrule(lr){15-18}
        \multicolumn{1}{l}{Setting} & Method & 1\% & 5\% & 10\% & 20\% & 1\% & 5\% & 10\% & 20\% & 1\% & 5\% & 10\% & 20\% & 1\% & 5\% & 10\% & 20\% \\
        \midrule
        \multirow{3}{*}{\shortstack[l]{Random\\No update}}
        & \mymethod{} (C) & 7.3 & 16.0 & 22.0 & 31.3 & 2.9 & 8.9 & 14.6 & 23.6 & 1.9 & 7.8 & 14.0 & 24.5 & 33.1 & 52.6 & 61.3 & 69.4 \\
        & \mymethod{} (U) & 6.8 & 15.5 & 22.5 & 32.9 & 2.1 & 8.1 & 14.5 & 25.8 & 1.6 & 7.2 & 13.3 & 24.4 & 31.6 & 54.6 & 62.8 & 70.8 \\
        & \mymethod{} (U*) & 6.9 & 15.6 & 22.6 & 32.9 & 2.0 & 8.2 & 14.5 & 25.7 & 1.7 & 7.3 & 13.4 & 24.5 & 32.0 & 54.1 & 63.3 & 70.8 \\
        \midrule
        \multirow{3}{*}{\shortstack[l]{Random\\Weighted avg}}
        & \mymethod{} (C) & 7.3 & 16.0 & 22.0 & 31.4 & 2.9 & 8.9 & 14.6 & 23.6 & 1.9 & 7.8 & 14.0 & 24.5 & 32.9 & 52.4 & 61.6 & 69.4 \\
        & \mymethod{} (U) & 7.2 & 15.8 & 21.9 & 31.2 & 2.9 & 8.9 & 14.4 & 23.6 & 1.8 & 7.7 & 13.8 & 24.2 & 32.3 & 54.0 & 61.6 & 69.9 \\
        & \mymethod{} (U*) & 7.3 & 16.0 & 22.0 & 31.4 & 2.9 & 8.9 & 14.6 & 23.6 & 1.9 & 7.8 & 14.0 & 24.5 & 32.8 & 54.0 & 61.8 & 69.8 \\
        \midrule
        \multirow{3}{*}{\shortstack[l]{Random\\Most similar}}
        & \mymethod{} (C) & 7.2 & 16.0 & 21.9 & 31.3 & 2.9 & 8.9 & 14.6 & 23.7 & 1.9 & 7.8 & 14.0 & 24.5 & 34.5 & 53.0 & 61.7 & 69.1 \\
        & \mymethod{} (U) & 7.1 & 15.7 & 22.6 & 32.6 & 2.7 & 9.0 & 15.3 & 26.2 & 1.8 & 7.4 & 13.6 & 24.4 & 31.1 & 53.2 & 62.7 & 70.0 \\
        & \mymethod{} (U*) & 7.2 & 15.9 & 22.6 & 32.6 & 2.7 & 9.0 & 15.3 & 26.3 & 1.9 & 7.6 & 13.7 & 24.6 & 29.9 & 53.5 & 62.9 & 70.3 \\
        \midrule
        \multirow{3}{*}{\shortstack[l]{Random\\Global mean}}
        & \mymethod{} (C) & 7.3 & 16.1 & 22.0 & 31.3 & 2.9 & 8.9 & 14.6 & 23.6 & 1.9 & 7.8 & 14.0 & 24.5 & 32.6 & 53.0 & 61.6 & 68.7 \\
        & \mymethod{} (U) & 7.2 & 15.8 & 21.8 & 31.2 & 2.9 & 8.8 & 14.5 & 23.6 & 1.8 & 7.7 & 13.8 & 24.2 & 32.7 & 54.2 & 61.5 & 68.5 \\
        & \mymethod{} (U*) & 7.3 & 16.0 & 22.0 & 31.3 & 2.9 & 8.9 & 14.5 & 23.6 & 1.9 & 7.8 & 14.0 & 24.5 & 33.5 & 54.0 & 61.6 & 68.4 \\
        \midrule
        \multirow{3}{*}{\shortstack[l]{Anchored\\No update}}
        & \mymethod{} (C) & 7.2 & 15.7 & 21.9 & 31.1 & 2.5 & 8.8 & 14.6 & 23.8 & 1.9 & 7.8 & 14.1 & 24.7 & 33.6 & 55.4 & 63.3 & 70.3 \\
        & \mymethod{} (U) & 7.2 & 15.7 & 22.5 & 32.5 & 2.4 & 8.8 & 15.3 & 26.5 & 1.9 & 7.8 & 14.0 & 25.1 & 30.8 & 56.7 & 64.7 & 71.6 \\
        & \mymethod{} (U*) & 7.2 & 15.8 & 22.5 & 32.4 & 2.4 & 8.9 & 15.3 & 26.5 & 1.9 & 7.8 & 14.1 & 25.2 & 32.1 & 56.6 & 64.3 & 71.2 \\
    \bottomrule
    \end{tabular}}
\end{table}

\newpage

\subsection{Results on applying embedding update rules to uniformly anchored node embeddings}\label{app:update_rules}

\begin{table}[h]
    \centering
    \caption{Influence spread (\%) achieved by \mymethod{} with uniformly anchored node embeddings and different post-training embedding update rules.}
    \label{tab:update_rules}
    \resizebox{\textwidth}{!}{
    \begin{tabular}{l|lcccccccccccccccc@{}}
    \toprule
        \multicolumn{1}{l}{} & & \multicolumn{4}{c}{Cora-ML (IC)} & \multicolumn{4}{c}{Network Science (IC)} & \multicolumn{4}{c}{Power Grid (IC)} & \multicolumn{4}{c}{Jazz (IC)} \\
        \cmidrule(lr){3-6} \cmidrule(lr){7-10} \cmidrule(lr){11-14} \cmidrule(lr){15-18}
        \multicolumn{1}{l}{Setting} & Method & 1\% & 5\% & 10\% & 20\% & 1\% & 5\% & 10\% & 20\% & 1\% & 5\% & 10\% & 20\% & 1\% & 5\% & 10\% & 20\% \\
        \midrule
        \multirow{3}{*}{\shortstack[l]{Anchored\\Weighted avg}}
        & \mymethod{} (C) & 14.1 & 32.9 & 39.9 & 54.7 & 9.5 & 23.6 & 36.8 & 50.3 & 8.4 & 26.7 & 40.4 & 57.7 & 11.1 & 27.1 & 37.2 & 50.1 \\
        & \mymethod{} (U) & 16.9 & 32.9 & 43.2 & 57.6 & 9.7 & 23.8 & 36.8 & 58.0 & 8.2 & 26.0 & 40.6 & 59.7 & 12.1 & 26.5 & 37.1 & 50.7 \\
        & \mymethod{} (U*) & 16.9 & 33.2 & 43.3 & 57.7 & 9.8 & 24.1 & 37.0 & 58.3 & 8.4 & 26.9 & 41.3 & 60.0 & 11.9 & 26.6 & 37.4 & 50.4 \\
        \midrule
        \multirow{3}{*}{\shortstack[l]{Anchored\\Most similar}}
        & \mymethod{} (C) & 14.3 & 33.2 & 39.8 & 54.5 & 9.5 & 24.0 & 36.9 & 50.1 & 8.4 & 26.6 & 40.4 & 57.5 & 10.5 & 27.8 & 36.6 & 50.2 \\
        & \mymethod{} (U) & 16.2 & 33.1 & 43.2 & 57.1 & 9.6 & 23.8 & 36.9 & 58.1 & 8.2 & 25.9 & 40.3 & 59.4 & 12.1 & 27.4 & 37.4 & 50.0 \\
        & \mymethod{} (U*) & 16.5 & 33.6 & 43.3 & 57.1 & 9.7 & 24.2 & 37.2 & 58.1 & 8.4 & 26.8 & 41.2 & 59.9 & 12.0 & 27.7 & 36.7 & 50.2 \\
        \midrule
        \multirow{3}{*}{\shortstack[l]{Anchored\\Global mean}}
        & \mymethod{} (C) & 14.2 & 33.2 & 39.9 & 54.6 & 9.5 & 24.1 & 36.8 & 50.3 & 8.5 & 26.6 & 40.4 & 57.5 & 10.1 & 27.0 & 36.8 & 50.0 \\
        & \mymethod{} (U) & 17.2 & 33.1 & 43.4 & 56.9 & 9.6 & 24.1 & 37.1 & 58.1 & 8.1 & 26.1 & 40.5 & 59.6 & 11.9 & 26.8 & 37.5 & 50.3 \\
        & \mymethod{} (U*) & 16.9 & 33.5 & 43.2 & 57.0 & 9.5 & 24.3 & 37.4 & 58.3 & 8.4 & 26.8 & 41.4 & 60.0 & 11.9 & 26.7 & 37.8 & 50.2 \\
        \midrule
        \multirow{3}{*}{\shortstack[l]{Anchored\\No update}}
        & \mymethod{} (C) & 14.2 & 33.2 & 39.8 & 54.7 & 9.5 & 24.0 & 36.6 & 50.3 & 8.5 & 26.7 & 40.5 & 57.6 & 10.4 & 27.2 & 36.7 & 50.0 \\
        & \mymethod{} (U) & 16.9 & 33.3 & 43.1 & 57.3 & 9.7 & 24.0 & 36.6 & 57.8 & 8.2 & 26.2 & 40.4 & 59.5 & 12.3 & 27.6 & 37.5 & 51.1 \\
        & \mymethod{} (U*) & 17.2 & 33.6 & 43.3 & 57.4 & 9.6 & 24.4 & 37.1 & 58.1 & 8.5 & 26.9 & 41.3 & 60.1 & 11.7 & 26.7 & 37.5 & 50.8 \\
    \bottomrule
    \end{tabular}}
    
    \resizebox{\textwidth}{!}{
    \begin{tabular}{l|lcccccccccccccccc@{}}
    \toprule
        \multicolumn{1}{l}{} & & \multicolumn{4}{c}{Cora-ML (LT)} & \multicolumn{4}{c}{Network Science (LT)} & \multicolumn{4}{c}{Power Grid (LT)} & \multicolumn{4}{c}{Jazz (LT)} \\
        \cmidrule(lr){3-6} \cmidrule(lr){7-10} \cmidrule(lr){11-14} \cmidrule(lr){15-18}
        \multicolumn{1}{l}{Setting} & Method & 1\% & 5\% & 10\% & 20\% & 1\% & 5\% & 10\% & 20\% & 1\% & 5\% & 10\% & 20\% & 1\% & 5\% & 10\% & 20\% \\
        \midrule
        \multirow{3}{*}{\shortstack[l]{Anchored\\Weighted avg}}
        & \mymethod{} (C) & 20.7 & 76.9 & 84.8 & 89.4 & 4.8 & 22.0 & 28.6 & 41.9 & 7.8 & 32.4 & 54.2 & 77.5 & 1.3 & 7.4 & 70.5 & 98.3 \\
        & \mymethod{} (U) & 17.9 & 80.4 & 93.1 & 98.8 & 7.1 & 26.2 & 42.2 & 65.7 & 7.5 & 30.9 & 58.1 & 88.4 & 2.0 & 30.4 & 52.4 & 100.0 \\
        & \mymethod{} (U*) & 19.8 & 81.3 & 93.4 & 98.7 & 7.0 & 26.4 & 42.2 & 66.0 & 7.8 & 32.5 & 59.5 & 88.7 & 2.0 & 26.8 & 60.1 & 100.0 \\
        \midrule
        \multirow{3}{*}{\shortstack[l]{Anchored\\Most similar}}
        & \mymethod{} (C) & 20.6 & 76.7 & 84.8 & 89.4 & 4.6 & 21.7 & 28.6 & 41.9 & 7.8 & 32.4 & 54.3 & 77.4 & 1.4 & 7.5 & 70.6 & 98.2 \\
        & \mymethod{} (U) & 17.2 & 80.4 & 92.7 & 98.8 & 7.1 & 25.4 & 42.1 & 66.9 & 7.6 & 30.7 & 58.3 & 88.3 & 1.9 & 30.4 & 50.7 & 100.0 \\
        & \mymethod{} (U*) & 18.1 & 80.1 & 92.5 & 98.9 & 7.2 & 25.6 & 41.9 & 66.9 & 7.9 & 32.1 & 59.2 & 88.5 & 1.9 & 26.9 & 59.5 & 100.0 \\
        \midrule
        \multirow{3}{*}{\shortstack[l]{Anchored\\Global mean}}
        & \mymethod{} (C) & 20.5 & 76.8 & 84.7 & 89.4 & 4.8 & 21.9 & 28.6 & 41.9 & 7.8 & 32.5 & 54.3 & 77.4 & 1.3 & 7.4 & 71.5 & 98.1 \\
        & \mymethod{} (U) & 18.3 & 80.2 & 93.2 & 98.8 & 6.8 & 25.8 & 42.4 & 66.6 & 7.6 & 30.7 & 58.0 & 88.4 & 2.0 & 30.6 & 54.6 & 100.0 \\
        & \mymethod{} (U*) & 18.7 & 80.2 & 93.5 & 98.8 & 7.2 & 26.2 & 42.3 & 66.8 & 7.8 & 32.5 & 59.0 & 88.8 & 2.0 & 21.8 & 59.3 & 100.0 \\
        \midrule
        \multirow{3}{*}{\shortstack[l]{Anchored\\No update}}
        & \mymethod{} (C) & 20.7 & 76.7 & 84.8 & 89.4 & 4.9 & 22.1 & 28.6 & 41.9 & 7.8 & 32.5 & 54.2 & 77.4 & 1.3 & 7.4 & 71.2 & 98.6 \\
        & \mymethod{} (U) & 18.9 & 80.0 & 93.3 & 98.8 & 6.8 & 26.5 & 41.7 & 66.4 & 7.6 & 31.0 & 58.4 & 88.1 & 2.0 & 30.4 & 48.0 & 100.0 \\
        & \mymethod{} (U*) & 18.0 & 80.3 & 93.0 & 98.9 & 7.0 & 26.3 & 41.6 & 66.6 & 7.8 & 32.4 & 59.5 & 88.7 & 2.0 & 22.5 & 58.5 & 100.0 \\
    \bottomrule
    \end{tabular}}
    
    \resizebox{\textwidth}{!}{
    \begin{tabular}{l|lcccccccccccccccc@{}}
    \toprule
        \multicolumn{1}{l}{} & & \multicolumn{4}{c}{Cora-ML (SIS)} & \multicolumn{4}{c}{Network Science (SIS)} & \multicolumn{4}{c}{Power Grid (SIS)} & \multicolumn{4}{c}{Jazz (SIS)} \\
        \cmidrule(lr){3-6} \cmidrule(lr){7-10} \cmidrule(lr){11-14} \cmidrule(lr){15-18}
        \multicolumn{1}{l}{Setting} & Method & 1\% & 5\% & 10\% & 20\% & 1\% & 5\% & 10\% & 20\% & 1\% & 5\% & 10\% & 20\% & 1\% & 5\% & 10\% & 20\% \\
        \midrule
        \multirow{3}{*}{\shortstack[l]{Anchored\\Weighted avg}}
        & \mymethod{} (C) & 7.3 & 15.8 & 21.9 & 31.0 & 2.5 & 8.8 & 14.6 & 23.8 & 1.9 & 7.8 & 14.1 & 24.7 & 33.7 & 55.0 & 63.0 & 70.5 \\
        & \mymethod{} (U) & 7.1 & 15.8 & 22.2 & 32.4 & 2.4 & 8.8 & 15.4 & 26.4 & 1.9 & 7.7 & 13.9 & 25.2 & 28.4 & 56.2 & 64.2 & 71.5 \\
        & \mymethod{} (U*) & 7.2 & 15.8 & 22.4 & 32.4 & 2.4 & 8.9 & 15.5 & 26.4 & 1.9 & 7.8 & 14.1 & 25.3 & 31.1 & 55.9 & 64.6 & 71.4 \\
        \midrule
        \multirow{3}{*}{\shortstack[l]{Anchored\\Most similar}}
        & \mymethod{} (C) & 7.3 & 15.8 & 21.9 & 31.1 & 2.6 & 8.7 & 14.7 & 23.8 & 1.9 & 7.8 & 14.1 & 24.7 & 33.3 & 55.1 & 62.7 & 70.1 \\
        & \mymethod{} (U) & 7.1 & 15.7 & 22.4 & 32.5 & 2.5 & 8.9 & 15.4 & 26.0 & 1.9 & 7.7 & 13.9 & 25.1 & 29.9 & 56.1 & 64.5 & 71.5 \\
        & \mymethod{} (U*) & 7.2 & 15.8 & 22.3 & 32.4 & 2.5 & 8.8 & 15.3 & 26.0 & 1.9 & 7.9 & 14.1 & 25.2 & 32.1 & 55.9 & 64.7 & 71.1 \\
        \midrule
        \multirow{3}{*}{\shortstack[l]{Anchored\\Global mean}}
        & \mymethod{} (C) & 7.3 & 15.8 & 21.9 & 31.1 & 2.5 & 8.8 & 14.6 & 23.8 & 1.9 & 7.8 & 14.1 & 24.7 & 34.7 & 55.7 & 63.3 & 70.5 \\
        & \mymethod{} (U) & 7.1 & 15.8 & 22.3 & 32.5 & 2.3 & 8.9 & 15.3 & 26.2 & 1.9 & 7.8 & 14.0 & 25.1 & 29.9 & 56.3 & 64.4 & 71.4 \\
        & \mymethod{} (U*) & 7.2 & 15.8 & 22.3 & 32.6 & 2.3 & 8.9 & 15.4 & 26.1 & 1.9 & 7.8 & 14.1 & 25.3 & 30.9 & 56.3 & 64.6 & 71.4 \\
        \midrule
        \multirow{3}{*}{\shortstack[l]{Anchored\\No update}}
        & \mymethod{} (C) & 7.2 & 15.7 & 21.9 & 31.1 & 2.5 & 8.8 & 14.6 & 23.8 & 1.9 & 7.8 & 14.1 & 24.7 & 33.6 & 55.4 & 63.3 & 70.3 \\
        & \mymethod{} (U) & 7.2 & 15.7 & 22.5 & 32.5 & 2.4 & 8.8 & 15.3 & 26.5 & 1.9 & 7.8 & 14.0 & 25.1 & 30.8 & 56.7 & 64.7 & 71.6 \\
        & \mymethod{} (U*) & 7.2 & 15.8 & 22.5 & 32.4 & 2.4 & 8.9 & 15.3 & 26.5 & 1.9 & 7.8 & 14.1 & 25.2 & 32.1 & 56.6 & 64.3 & 71.2 \\
    \bottomrule
    \end{tabular}}
\end{table}

\newpage

\subsection{Influence maximization performance of \mymethod{} with different node embedding dimensions}\label{app:emb_dim_2_vs_8}

\begin{table}[h]
    \centering
    \caption{Influence spread (\%) achieved by \mymethod{} with 2D uniformly anchored node embeddings.}
    \label{tab:emb_dim_2_vs_8}
    \resizebox{\textwidth}{!}{
    \begin{tabular}{c|lcccccccccccccccc@{}}
    \toprule
        \multicolumn{1}{l}{} & & \multicolumn{4}{c}{Cora-ML (IC)} & \multicolumn{4}{c}{Network Science (IC)} & \multicolumn{4}{c}{Power Grid (IC)} & \multicolumn{4}{c}{Jazz (IC)} \\
        \cmidrule(lr){3-6} \cmidrule(lr){7-10} \cmidrule(lr){11-14} \cmidrule(lr){15-18}
        \multicolumn{1}{c}{Embedding} & Method & 1\% & 5\% & 10\% & 20\% & 1\% & 5\% & 10\% & 20\% & 1\% & 5\% & 10\% & 20\% & 1\% & 5\% & 10\% & 20\% \\
        \midrule
        \multirow{3}{*}{$d = 2$}
        & \mymethod{} (C) & 13.9 & 32.8 & 39.4 & 54.5 & 9.2 & 24.0 & 36.9 & 50.2 & 8.3 & 26.3 & 40.0 & 57.3 & 11.0 & 27.0 & 36.3 & 49.5 \\
        & \mymethod{} (U) & 16.9 & 32.7 & 42.6 & 56.9 & 9.4 & 24.1 & 37.2 & 57.6 & 8.0 & 25.9 & 40.2 & 59.3 & 11.2 & 27.0 & 36.8 & 51.0 \\
        & \mymethod{} (U*) & 16.9 & 33.2 & 42.6 & 56.9 & 9.4 & 24.3 & 37.4 & 57.8 & 8.3 & 26.7 & 40.9 & 59.8 & 12.0 & 27.8 & 36.9 & 50.9 \\
        \midrule
        \multirow{3}{*}{$d = 8$}
        & \mymethod{} (C) & 14.2 & 33.2 & 39.8 & 54.7 & 9.5 & 24.0 & 36.6 & 50.3 & 8.5 & 26.7 & 40.5 & 57.6 & 10.4 & 27.2 & 36.7 & 50.0 \\
        & \mymethod{} (U) & 16.9 & 33.3 & 43.1 & 57.3 & 9.7 & 24.0 & 36.6 & 57.8 & 8.2 & 26.2 & 40.4 & 59.5 & 12.3 & 27.6 & 37.5 & 51.1 \\
        & \mymethod{} (U*) & 17.2 & 33.6 & 43.3 & 57.4 & 9.6 & 24.4 & 37.1 & 58.1 & 8.5 & 26.9 & 41.3 & 60.1 & 11.7 & 26.7 & 37.5 & 50.8 \\
    \bottomrule
    \end{tabular}}
    
    \resizebox{\textwidth}{!}{
    \begin{tabular}{c|lcccccccccccccccc@{}}
    \toprule
        \multicolumn{1}{l}{} & & \multicolumn{4}{c}{Cora-ML (LT)} & \multicolumn{4}{c}{Network Science (LT)} & \multicolumn{4}{c}{Power Grid (LT)} & \multicolumn{4}{c}{Jazz (LT)} \\
        \cmidrule(lr){3-6} \cmidrule(lr){7-10} \cmidrule(lr){11-14} \cmidrule(lr){15-18}
        \multicolumn{1}{c}{Embedding} & Method & 1\% & 5\% & 10\% & 20\% & 1\% & 5\% & 10\% & 20\% & 1\% & 5\% & 10\% & 20\% & 1\% & 5\% & 10\% & 20\% \\
        \midrule
        \multirow{3}{*}{$d = 2$}
        & \mymethod{} (C) & 19.3 & 75.1 & 84.8 & 89.3 & 4.8 & 21.8 & 28.6 & 41.9 & 7.7 & 32.3 & 54.3 & 77.4 & 1.3 & 7.5 & 71.9 & 95.4 \\
        & \mymethod{} (U) & 17.8 & 77.6 & 92.7 & 98.5 & 6.9 & 26.0 & 41.3 & 65.7 & 7.5 & 30.7 & 58.6 & 88.4 & 1.8 & 30.4 & 59.9 & 99.8 \\
        & \mymethod{} (U*) & 18.5 & 77.2 & 92.8 & 98.5 & 7.0 & 26.0 & 41.6 & 65.5 & 7.8 & 32.2 & 59.5 & 88.7 & 1.8 & 26.0 & 60.9 & 99.9 \\
        \midrule
        \multirow{3}{*}{$d = 8$}
        & \mymethod{} (C) & 20.7 & 76.7 & 84.8 & 89.4 & 4.9 & 22.1 & 28.6 & 41.9 & 7.8 & 32.5 & 54.2 & 77.4 & 1.3 & 7.4 & 71.2 & 98.6 \\
        & \mymethod{} (U) & 18.9 & 80.0 & 93.3 & 98.8 & 6.8 & 26.5 & 41.7 & 66.4 & 7.6 & 31.0 & 58.4 & 88.1 & 2.0 & 30.4 & 48.0 & 100.0 \\
        & \mymethod{} (U*) & 18.0 & 80.3 & 93.0 & 98.9 & 7.0 & 26.3 & 41.6 & 66.6 & 7.8 & 32.4 & 59.5 & 88.7 & 2.0 & 22.5 & 58.5 & 100.0 \\
    \bottomrule
    \end{tabular}}
    
    \resizebox{\textwidth}{!}{
    \begin{tabular}{c|lcccccccccccccccc@{}}
    \toprule
        \multicolumn{1}{l}{} & & \multicolumn{4}{c}{Cora-ML (SIS)} & \multicolumn{4}{c}{Network Science (SIS)} & \multicolumn{4}{c}{Power Grid (SIS)} & \multicolumn{4}{c}{Jazz (SIS)} \\
        \cmidrule(lr){3-6} \cmidrule(lr){7-10} \cmidrule(lr){11-14} \cmidrule(lr){15-18}
        \multicolumn{1}{c}{Embedding} & Method & 1\% & 5\% & 10\% & 20\% & 1\% & 5\% & 10\% & 20\% & 1\% & 5\% & 10\% & 20\% & 1\% & 5\% & 10\% & 20\% \\
        \midrule
        \multirow{3}{*}{$d = 2$}
        & \mymethod{} (C) & 7.3 & 15.8 & 21.9 & 31.0 & 2.6 & 8.8 & 14.7 & 23.8 & 1.9 & 7.9 & 14.1 & 24.7 & 34.4 & 54.6 & 63.3 & 69.9 \\
        & \mymethod{} (U) & 7.2 & 15.8 & 22.4 & 32.3 & 2.5 & 8.7 & 15.4 & 26.2 & 1.8 & 7.7 & 14.0 & 25.1 & 34.7 & 56.2 & 64.5 & 70.6 \\
        & \mymethod{} (U*) & 7.3 & 16.0 & 22.3 & 32.3 & 2.5 & 8.8 & 15.3 & 26.2 & 1.9 & 7.9 & 14.2 & 25.2 & 34.5 & 55.9 & 64.4 & 71.1 \\
        \midrule
        \multirow{3}{*}{$d = 8$}
        & \mymethod{} (C) & 7.2 & 15.7 & 21.9 & 31.1 & 2.5 & 8.8 & 14.6 & 23.8 & 1.9 & 7.8 & 14.1 & 24.7 & 33.6 & 55.4 & 63.3 & 70.3 \\
        & \mymethod{} (U) & 7.2 & 15.7 & 22.5 & 32.5 & 2.4 & 8.8 & 15.3 & 26.5 & 1.9 & 7.8 & 14.0 & 25.1 & 30.8 & 56.7 & 64.7 & 71.6 \\
        & \mymethod{} (U*) & 7.2 & 15.8 & 22.5 & 32.4 & 2.4 & 8.9 & 15.3 & 26.5 & 1.9 & 7.8 & 14.1 & 25.2 & 32.1 & 56.6 & 64.3 & 71.2 \\
    \bottomrule
    \end{tabular}}
\end{table}

\newpage

\subsection{Architectural robustness}\label{app:binary_vs_8}

\begin{table}[h]
    \centering
    \caption{Influence spread (\%) achieved by \mymethod{} with binary seed vector input.}
    \label{tab:emb_dim_binary_vs_8}
    \resizebox{\textwidth}{!}{
    \begin{tabular}{c|lcccccccccccccccc@{}}
    \toprule
        \multicolumn{1}{l}{} & & \multicolumn{4}{c}{Cora-ML (IC)} & \multicolumn{4}{c}{Network Science (IC)} & \multicolumn{4}{c}{Power Grid (IC)} & \multicolumn{4}{c}{Jazz (IC)} \\
        \cmidrule(lr){3-6} \cmidrule(lr){7-10} \cmidrule(lr){11-14} \cmidrule(lr){15-18}
        \multicolumn{1}{c}{Embedding} & Method & 1\% & 5\% & 10\% & 20\% & 1\% & 5\% & 10\% & 20\% & 1\% & 5\% & 10\% & 20\% & 1\% & 5\% & 10\% & 20\% \\
        \midrule
        \multirow{3}{*}{Binary}
        & \mymethod{} (C) & 13.9 & 31.7 & 39.7 & 54.3 & 9.3 & 24.0 & 36.9 & 50.2 & 8.1 & 26.3 & 40.0 & 57.4 & 10.0 & 27.4 & 36.2 & 50.3 \\
        & \mymethod{} (U) & 16.9 & 31.8 & 43.2 & 56.1 & 9.6 & 24.1 & 37.3 & 57.9 & 8.2 & 26.0 & 40.2 & 59.2 & 11.1 & \textbf{27.7} & \textbf{37.9} & 50.5 \\
        & \mymethod{} (U*) & 17.0 & 31.8 & 42.9 & 56.1 & 9.5 & 24.3 & \textbf{37.4} & \textbf{58.3} & 8.3 & 26.6 & 41.1 & 59.8 & 11.7 & 27.6 & 38.2 & 50.3 \\
        \midrule
        \multirow{3}{*}{$d = 8$}
        & \mymethod{} (C) & 14.2 & 33.2 & 39.8 & 54.7 & 9.5 & 24.0 & 36.6 & 50.3 & \textbf{8.5} & 26.7 & 40.5 & 57.6 & 10.4 & 27.2 & 36.7 & 50.0 \\
        & \mymethod{} (U) & 16.9 & 33.3 & 43.1 & 57.3 & \textbf{9.7} & 24.0 & 36.6 & 57.8 & 8.2 & 26.2 & 40.4 & 59.5 & \textbf{12.3} & 27.6 & 37.5 & \textbf{51.1} \\
        & \mymethod{} (U*) & \textbf{17.2} & \textbf{33.6} & \textbf{43.3} & \textbf{57.4} & 9.6 & \textbf{24.4} & 37.1 & 58.1 & \textbf{8.5} & \textbf{26.9} & \textbf{41.3} & \textbf{60.1} & 11.7 & 26.7 & 37.5 & 50.8 \\
    \bottomrule
    \end{tabular}}
    
    \resizebox{\textwidth}{!}{
    \begin{tabular}{c|lcccccccccccccccc@{}}
    \toprule
        \multicolumn{1}{l}{} & & \multicolumn{4}{c}{Cora-ML (LT)} & \multicolumn{4}{c}{Network Science (LT)} & \multicolumn{4}{c}{Power Grid (LT)} & \multicolumn{4}{c}{Jazz (LT)} \\
        \cmidrule(lr){3-6} \cmidrule(lr){7-10} \cmidrule(lr){11-14} \cmidrule(lr){15-18}
        \multicolumn{1}{c}{Embedding} & Method & 1\% & 5\% & 10\% & 20\% & 1\% & 5\% & 10\% & 20\% & 1\% & 5\% & 10\% & 20\% & 1\% & 5\% & 10\% & 20\% \\
        \midrule
        \multirow{3}{*}{Binary}
        & \mymethod{} (C) & 20.3 & 76.7 & 84.8 & 89.4 & 5.2 & 21.6 & 28.6 & 41.9 & 7.6 & 32.0 & 54.3 & 77.5 & 1.4 & 7.4 & \textbf{71.8} & 98.4 \\
        & \mymethod{} (U) & 17.7 & 79.9 & 92.8 & 98.6 & 6.8 & 26.0 & 41.5 & 64.6 & 7.3 & 30.2 & 58.2 & 88.4 & \textbf{2.0} & \textbf{30.5} & 49.5 & 99.9 \\
        & \mymethod{} (U*) & 18.2 & 80.2 & 92.8 & 98.7 & 7.0 & 26.1 & 41.5 & 64.5 & 7.7 & 31.7 & 59.1 & 88.4 & \textbf{2.0} & 26.4 & 56.9 & 99.9 \\
        \midrule
        \multirow{3}{*}{$d = 8$}
        & \mymethod{} (C) & \textbf{20.7} & 76.7 & 84.8 & 89.4 & 4.9 & 22.1 & 28.6 & 41.9 & \textbf{7.8} & \textbf{32.5} & 54.2 & 77.4 & 1.3 & 7.4 & 71.2 & 98.6 \\
        & \mymethod{} (U) & 18.9 & 80.0 & \textbf{93.3} & 98.8 & 6.8 & \textbf{26.5} & \textbf{41.7} & 66.4 & 7.6 & 31.0 & 58.4 & 88.1 & \textbf{2.0} & 30.4 & 48.0 & \textbf{100.0} \\
        & \mymethod{} (U*) & 18.0 & \textbf{80.3} & 93.0 & \textbf{98.9} & \textbf{7.0} & 26.3 & 41.6 & \textbf{66.6} & \textbf{7.8} & 32.4 & \textbf{59.5} & \textbf{88.7} & \textbf{2.0} & 22.5 & 58.5 & \textbf{100.0} \\
    \bottomrule
    \end{tabular}}
    
    \resizebox{\textwidth}{!}{
    \begin{tabular}{c|lcccccccccccccccc@{}}
    \toprule
        \multicolumn{1}{l}{} & & \multicolumn{4}{c}{Cora-ML (SIS)} & \multicolumn{4}{c}{Network Science (SIS)} & \multicolumn{4}{c}{Power Grid (SIS)} & \multicolumn{4}{c}{Jazz (SIS)} \\
        \cmidrule(lr){3-6} \cmidrule(lr){7-10} \cmidrule(lr){11-14} \cmidrule(lr){15-18}
        \multicolumn{1}{c}{Embedding} & Method & 1\% & 5\% & 10\% & 20\% & 1\% & 5\% & 10\% & 20\% & 1\% & 5\% & 10\% & 20\% & 1\% & 5\% & 10\% & 20\% \\
        \midrule
        \multirow{3}{*}{Binary}
        & \mymethod{} (C) & \textbf{7.4} & \textbf{15.8} & 21.9 & 31.2 & \textbf{2.5} & \textbf{8.9} & 14.7 & 23.8 & \textbf{1.9} & \textbf{7.8} & \textbf{14.1} & 24.7 & 35.1 & 54.8 & 63.4 & 70.5 \\
        & \mymethod{} (U) & 7.2 & 15.7 & 22.5 & \textbf{32.6} & \textbf{2.5} & 8.6 & 14.8 & 26.4 & \textbf{1.9} & 7.7 & 14.0 & 25.1 & \textbf{35.3} & 56.4 & 64.6 & 71.5 \\
        & \mymethod{} (U*) & 7.2 & 1\textbf{5.8} & \textbf{22.6} & 32.4 & \textbf{2.5} & 8.5 & 14.8 & 26.3 & \textbf{1.9} & \textbf{7.8} & \textbf{14.1} & \textbf{25.3} & \textbf{35.3} & \textbf{56.7} & 64.6 & 71.5 \\
        \midrule
        \multirow{3}{*}{$d = 8$}
        & \mymethod{} (C) & 7.2 & 15.7 & 21.9 & 31.1 & \textbf{2.5} & 8.8 & 14.6 & 23.8 & \textbf{1.9} & \textbf{7.8} & \textbf{14.1} & 24.7 & 33.6 & 55.4 & 63.3 & 70.3 \\
        & \mymethod{} (U) & 7.2 & 15.7 & 22.5 & 32.5 & 2.4 & 8.8 & \textbf{15.3} & \textbf{26.5} & \textbf{1.9} & \textbf{7.8} & 14.0 & 25.1 & 30.8 & \textbf{56.7} & \textbf{64.7} & \textbf{71.6} \\
        & \mymethod{} (U*) & 7.2 & \textbf{15.8} & 22.5 & 32.4 & 2.4 & \textbf{8.9} & \textbf{15.3} & \textbf{26.5} & \textbf{1.9} & \textbf{7.8} & \textbf{14.1} & 25.2 & 32.1 & 56.6 & 64.3 & 71.2 \\
    \bottomrule
    \end{tabular}}
\end{table}

Finally, we evaluate an extreme lightweight variant where each node is represented using only its binary seed indicator (i.e., same input design as DeepIM). As reported in Table~\ref{tab:emb_dim_binary_vs_8}, although learnable embeddings boost IM performance in most cases, this simplified representation remains surprisingly competitive, suggesting that most of the predictive power arises from topology-aware and diffusion-dependent message passing rather than from high-dimensional node embeddings. This provides further evidence to support our core design philosophy that a simple, lightweight neural surrogate model is sufficient to guide the search for high-quality seed sets.

\end{document}